\documentclass[lettersize,journal]{IEEEtran}
\usepackage{colortbl}
\usepackage{amsmath,amsfonts}
\usepackage{algorithmic}
\usepackage{algorithm}
\usepackage{array}
\usepackage[caption=false,font=normalsize,labelfont=sf,textfont=sf]{subfig}
\usepackage{textcomp}
\usepackage{stfloats}
\usepackage{url}
\usepackage{verbatim}
\usepackage{graphicx}
\usepackage{cite}
\usepackage[numbers]{natbib}

\usepackage{multirow}
\usepackage[normalem]{ulem}
\useunder{\uline}{\ul}{}
\usepackage{subfig}
\usepackage{color}
\usepackage{xcolor}
\usepackage{arydshln}
\usepackage{amsmath,bm}
\usepackage{picinpar}
\usepackage{graphicx}
\usepackage[switch]{lineno}

\makeatletter
\long\def\figwindownonum[#1,#2,#3,#4] {
	\begin{window}[#1,#2,{#3},{\centering#4\par}] }
	\def\endfigwindownonum{\end{window}}
\makeatother

\begin{document}

\title{Towards Fast and Accurate Image-Text Retrieval with Self-Supervised Fine-Grained Alignment}


\author{Jiamin Zhuang, Jing Yu$^{*}$, Yang Ding, Xiangyan Qu, Yue Hu
\thanks{This work was supported by the National Natural Science Foundation of China (Grant No. 62006222) and the Youth Innovation Promotion Association of CAS (Grant No. 2021153).}    
\thanks{Jiamin Zhuang, Jing Yu, Yang Ding, Xiangyan Qu and Yue Hu are with the Institute of Information Engineering, Chinese Academy of Sciences, China, and the School of Cyber Security, University of Chinese Academy of Sciences, China. (e-mail: zhuangjiamin@iie.ac.cn; yujing02@iie.ac.cn; dingyang@iie.ac.cn; quxiangyan@iie.ac.cn; huyue@iie.ac.cn)}
\thanks{Corresponding author: Jing Yu (e-mail: yujing02@iie.ac.cn)}
}

\markboth{Journal of \LaTeX\ Class Files,~Vol.~14, No.~8, August~2021}%
{Shell \MakeLowercase{\textit{et al.}}: A Sample Article Using IEEEtran.cls for IEEE Journals}


\maketitle

\begin{abstract}
Image-text retrieval requires the system to bridge the heterogenous gap between vision and language for accurate retrieval while keeping the network lightweight-enough for efficient retrieval. Existing trade-off solutions mainly study from the view of incorporating cross-modal interactions with the independent-embedding framework or leveraging stronger pre-trained encoders, which still demand  time-consuming similarity measurement or heavyweight model structure in the retrieval stage. In this work, we propose an image-text alignment module SelfAlign on top of the independent-embedding framework, which improves the retrieval accuracy  while maintains the retrieval efficiency without extra supervision. SelfAlign contains two collaborative sub-modules that force image-text alignment at both concept level and context level by self-supervised contrastive learning. It doesn’t require cross-modal embedding interactions \textcolor{black}{during} training while maintaining independent image and text encoders \textcolor{black}{during} retrieval. With comparable time cost, SelfAlign consistently boosts the accuracy of state-of-the-art non-pre-training independent-embedding models respectively by 9.1\%, 4.2\% and 6.6\% in terms of R@sum score on Flickr30K, MS-COCO 1K and MS-COCO 5K datasets. The retrieval accuracy also outperforms most existing interactive-embedding models with  orders of magnitude decrease in retrieval time. The source code is available at:
\url{https://github.com/Zjamie813/SelfAlign}.
\end{abstract}


\begin{figure}  
\centering  
\includegraphics[width=3.5in]{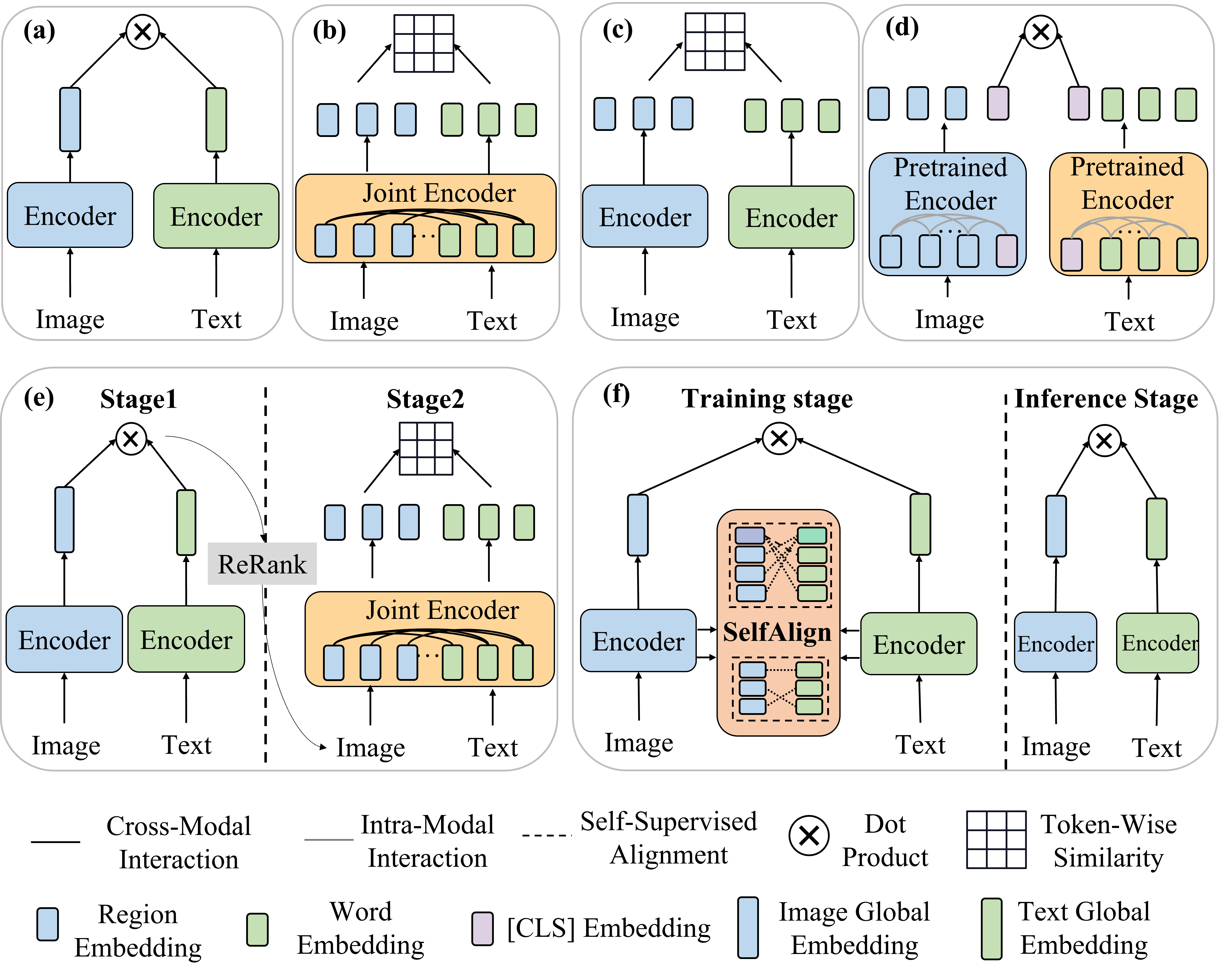}  
\caption{Illustration of different image-text retrieval approaches. (a) Independent-embedding approach. (b) Interactive-embedding approach. \textcolor{black}{(c) Late-interaction approach. (d) Intra-interactive embedding approach. (e) Two-stage approach.} (f) Independent-embedding approach with SelfAlign.}  
\label{Fig:intro} 
\vspace{-0.5cm}
\end{figure}

\section{Introduction}
\IEEEPARstart{I}{mage-text} retrieval (ITR) is a long-standing task that requires an AI agent to retrieve semantically relevant images given a text query and vice versa. The key challenge of ITR is to bridge the heterogeneous gap between low-level visual appearance and high-level abstract language and align their representations. It is also a fundamental problem for a series of vision and language tasks
\cite{liu2019improving, antol2015vqa, ku2020room}. In real-world scenarios, besides effective cross-modal alignment for accurate retrieval, the retrieval system also strives to make real-time retrieval possible with low latency. Therefore, how to balance the accuracy and efficiency becomes a key challenge for large-scale image-text retrieval.   

Most of the previous works make much effort on either retrieval efficiency or retrieval accuracy. Early \textit{independent-embedding approaches} \cite{faghri2017vse++, li2019visual, qu2020context} (Figure \ref{Fig:intro}(a)) encode each image and text independently into global embeddings. Then image-text similarity is computed by directly measuring the distance between their global embeddings in a common semantic space. Since there are no interactions between texts and images, independent-embedding approaches \textcolor{black}{allow} offline data embedding extraction and linear computational complexity \cite{liu2021inflate} for online retrieval. Hence such approaches are widely applied in real-world large-scale retrieval. However, \textcolor{black}{their retrieval accuracy is not satisfactory since such a global embedding alignment strategy cannot guarantee fine-grained content alignment.} To alleviate this problem, \textit{interactive-embedding approaches} \cite{lee2018stacked, chen2020imram,qu2021dynamic} (Figure \ref{Fig:intro}(b)) \textcolor{black}{are proposed for fine-grained image-text retrieval by aligning visual objects in an image with words in a text by cross-modal attention mechanism. However, for each query, all the retrieved samples need complex attention computation to encode their embeddings, which is  quite time-consuming and not scalable to large-scale online retrieval scenarios.} \textit{How to leverage the advantages of independent-embedding approaches and interactive-embedding approaches to achieve both high accuracy and practical efficiency becomes an \textcolor{black}{essential problem}.}

Current progress \cite{lu2021visualsparta,geigle2021retrieve, CLIP} aims to introduce computation-efficient interactions into the independent-embedding framework. The typical solutions can be divided into three types:  measuring fine-grained word-object similarities instead of global embedding similarities \textcolor{black}{(a.k.a \textit{late-interaction approach} as shown in Figure \ref{Fig:intro} (c))} \cite{lu2021visualsparta,liu2021inflate}, adopting independent-embedding approaches for coarse retrieval first and then using interactive-embedding approaches for finer retrieval \textcolor{black}{ (a.k.a. \textit{two-stage approach} as shown in Figure \ref{Fig:intro} (\textcolor{black}{e}))} \cite{geigle2021retrieve, miech2021thinking}, and  exploiting stronger intra-modal interactive encoder instead of \textcolor{black}{the} time-consuming cross-modal interactive encoder \textcolor{black}{(a.k.a. \textit{intra-interactive embedding approach} as shown in Figure \ref{Fig:intro} (\textcolor{black}{d})) }\cite{ALIGN,CLIP}. However, compared with independent-embedding approaches, these trade-off solutions still demand extra time cost in the retrieval stage due to complex similarity measurement, cross-modal embedding interactions, or heavyweight encoder structure. 
\textcolor{black}{They sacrifice the retrieval efficiency for the benefits of fine-grained feature learning.}

\textcolor{black}{In this paper, to enable fine-grained image-text \textcolor{black}{alignment} for accurate retrieval while maintain high retrieval efficiency as the independent-embedding models, we propose a novel trade-off strategy to learn fine-grained image-text alignment by contrastive-based embedding mapping. The advantage of contrastive-based embedding mapping is that it does not require cross-modal fusion during the inference stage. In our approach, we achieve embedding alignment between images and texts by a new module, named as SelfAlign. Based on the backbone of independent-embedding models, SelfAlign aims to align image and text embeddings from local to global by multi-level self-supervised contrastive learning. In this way, independent-embedding models injected with SelfAlign enhance original global embedding with more accurate fine-grained semantic alignment across different modalities. In the inference stage, the baseline model without the SelfAlign module conducts \textcolor{black}{image and text} encoding independently while maintains fine-grained embedding alignment ability. Therefore, SelfAlign enables independent-embedding models to achieve superior retrieval accuracy without sacrificing efficiency.}

\textcolor{black}{Specifically, SelfAlign is designed to explore fine-grained correspondence via mining rich visual and textual semantic content in different layers of the independent-embedding models. There are two sub-modules in SelfAlign responsible for semantic alignment at concept level and context level: (1) To capture the pair-wise correspondence among visual region and textual words, Local Concept Alignment (LCA) sub-module is first proposed to learn the concept-level alignment in the lower layer. (2) Since the semantically similar concepts have different semantics in different contexts, Context Relation Alignment (CRA) sub-module is further proposed to be injected into higher embedding layer to achieve context-level alignment. As a result, independent-embedding models with SelfAlign learn fine-grained alignment between images and texts from local semantics to global semantics progressively.}

The main contributions are summarized as follows: \textcolor{black}{(1) We introduce a novel trade-off strategy for image-text retrieval to learn fine-grained alignment by contrastive-based embedding mapping. The contrastive-based embedding mapping aligns the fine-grained image and text embeddings via cross-modal contrastive learning during the training stage without requiring cross-modal fusion during the inference stage. Thus, our approach has the benefits of both the interactive-embedding models and the independent-embedding models, i.e., enhancing fine-grained alignment learning while preserving the independent-embedding framework for efficient retrieval. (2) We propose a fine-grained image-text alignment module SelfAlign to achieve the trade-off strategy. Two sub-modules of SelfAlign conduct concept alignment and context alignment via cluster-based contrastive learning and global-to-local contrastive learning. Therefore, SelfAlign equips the global embedding of the independent-embedding models with multi-level fine-grained alignment to improve the retrieval accuracy without extra supervision.  (3) SelfAlign is a generic module that can be injected into various independent-embedding models. We incorporate SelfAlign with two representative independent-embedding models. Experimental results show that SelfAlign consistently boosts the accuracy of state-of-the-art independent-embedding models respectively by 9.1\%, 4.2\% and 6.6\% in terms of R@sum on Flickr30K, MS-COCO 1K and MS-COCO 5K. The performance also outperforms most existing interactive-embedding models with orders of magnitude decrease of retrieval time.}

\section{Related Works}
\subsection{Image-Text Retrieval}
Existing works can be categorized into two types: the independent-embedding approaches and the interactive-embedding approaches. The former approaches \cite{frome2013devise,faghri2017vse++,li2019visual,wu2019learning, qu2020context} aim to project the images and texts into a common semantic space, so image-text pairs can be compared directly via simple distance \textcolor{black}{metrics}. The architecture of mainstream independent-embedding approaches is \textcolor{black}{an} independent-embedding learning structure consisting of an image encoder and a text encoder, and they adopt ranking loss \cite{faghri2017vse++} for metric learning. Though these approaches have achieved some promising performance, they are still limited since they cannot conduct interactive encoding process and thus fail to provide fine-grained alignment between images and texts.

Interactive-embedding approaches \cite{karpathy2015deep,lee2018stacked,liu2020graph,chen2020imram,ji2021step,qu2021dynamic} aim to learn fine-grained image-text matching by complex object and word interactions with cross-modal attention mechanism. \citet{lee2018stacked} compute the similarities between regions and words, and only \textcolor{black}{count} the region-word pairs with high relevance. Some works \cite{liu2020graph, chen2020imram, ji2021step} propose hierarchical interaction methods for progressively extracting the complicated correspondence. Recently, \citet{qu2021dynamic} propose a dynamic router with the capability to choose the different interactive mode for each image-text pair and achieves state-of-the-art performance. Nevertheless, the quadratic computational complexity takes unavoidable computational cost for retrieval. In this paper, we propose a model-agnostic module with multi-level self-supervised learning strategy for independent-embedding models to learn the fine-grained semantic correspondences, instead of time-consuming attention mechanisms. Thus, the independent-embedding backbones achieve superior retrieval accuracy without sacrificing efficiency.

\subsection{Trade-Off Image-Text Retrieval Models}
To strike a balance between retrieval efficiency and accuracy, there are almost three \textcolor{black}{types of} approaches that have been proposed recently: late-interaction approaches \cite{lu2021visualsparta, liu2021inflate}, two-stage approaches \cite{geigle2021retrieve, miech2021thinking}, and intra-interactive embedding approaches \cite{sun2021lightningdot, ALIGN}. The late-interaction approaches retain the independent encoding architecture and perform lightweight token-wise interactions only in the late scoring stage \cite{lu2021visualsparta, liu2021inflate}. Their retrieval speed is still slower than independent-embedding methods since the independent-embedding methods only require the global embedding for similarity computation. Secondly, two-stage approaches \cite{geigle2021retrieve, miech2021thinking} first adopt independent-embedding models for coarse-level retrieval and then utilize interactive-embedding models for finer retrieval to trade-off between efficiency and accuracy. But they are still slower than independent-embedding models due to the existence of the re-rank stage. Lastly, the intra-interactive embedding approaches \cite{sun2021lightningdot, ALIGN} take two independent encoders but they require large-scale image-text pairs for training and stack a few Transformer blocks \cite{transformer} to build stronger encoders. For example, ALIGN \cite{ALIGN} leverages two transformers encoders with 400M parameters and 1.8B image-text pairs for training. Though they achieve surprising performance, the inevitable huge computation cost in training and deploying such massive-scale models limit their development. Contrast to these three types of trade-off methods, without sacrificing the retrieval time or requiring extra training data, our proposed module improves the accuracy by enhancing image and text representations of fine-grained information.

\subsection{Self-\textcolor{black}{Supervised} Contrastive Learning}
Self-supervised learning \cite{simclr, DBLP:conf/icml/TianCG21,DBLP:conf/cvpr/WuXYL18} aims at learning features without manual annotations. Recent approaches based on contrastive learning have achieved remarkable progress in visual domain. Current contrastive learning can be divided into two groups: individual-based contrastive learning \cite{simclr,moco} and cluster-based contrastive learning \cite{caron2020unsupervised, DBLP:conf/eccv/CaronBJD18}. Individual-based contrastive learning \cite{simclr,moco} considers each image in a dataset as its own class \cite{caron2020unsupervised}, and brings the embedding of different views from the same image closer and push embeddings from different images far apart using instance-level contrastive loss. This approach introduces the individual-level discrimination but requires a large batch size for negatives storage. Cluster-based contrastive learning \cite{caron2020unsupervised, DBLP:conf/eccv/CaronBJD18} encourages the image embeddings to be closer to their assigned prototypes obtained by clustering algorithm, and far from negative prototypes. This approach introduces the group-level discrimination between instances. However, current works only construct image-level representation learning and lack of fine-grained information learning such as the object information in images.

To learn local information, some researchers \cite{wang2021dense, li2021efficient} construct local level contrastive learning to learn visual pixel-level semantic information. Moreover, some works \cite{hjelm2018learning, bachman2019learning} maximize global-local mutual information and aim to learn the shared context information across patches/tokens. Here, we aim to learn the fine-grained correspondences between images and texts without detail annotations, and our word-object alignment learning and context-level alignment learning were inspired by these local contrastive learning works. Differently, the contrastive reasoning should be performed across modalities instead of cross-image views, and thus the learning process is not fully symmetric since the local semantic information involved in image-text pairs is not exactly equivalent to that of two augmentations of an image.

\section{Methodology}
\begin{figure*}  
\centering  
\includegraphics[width=6in]{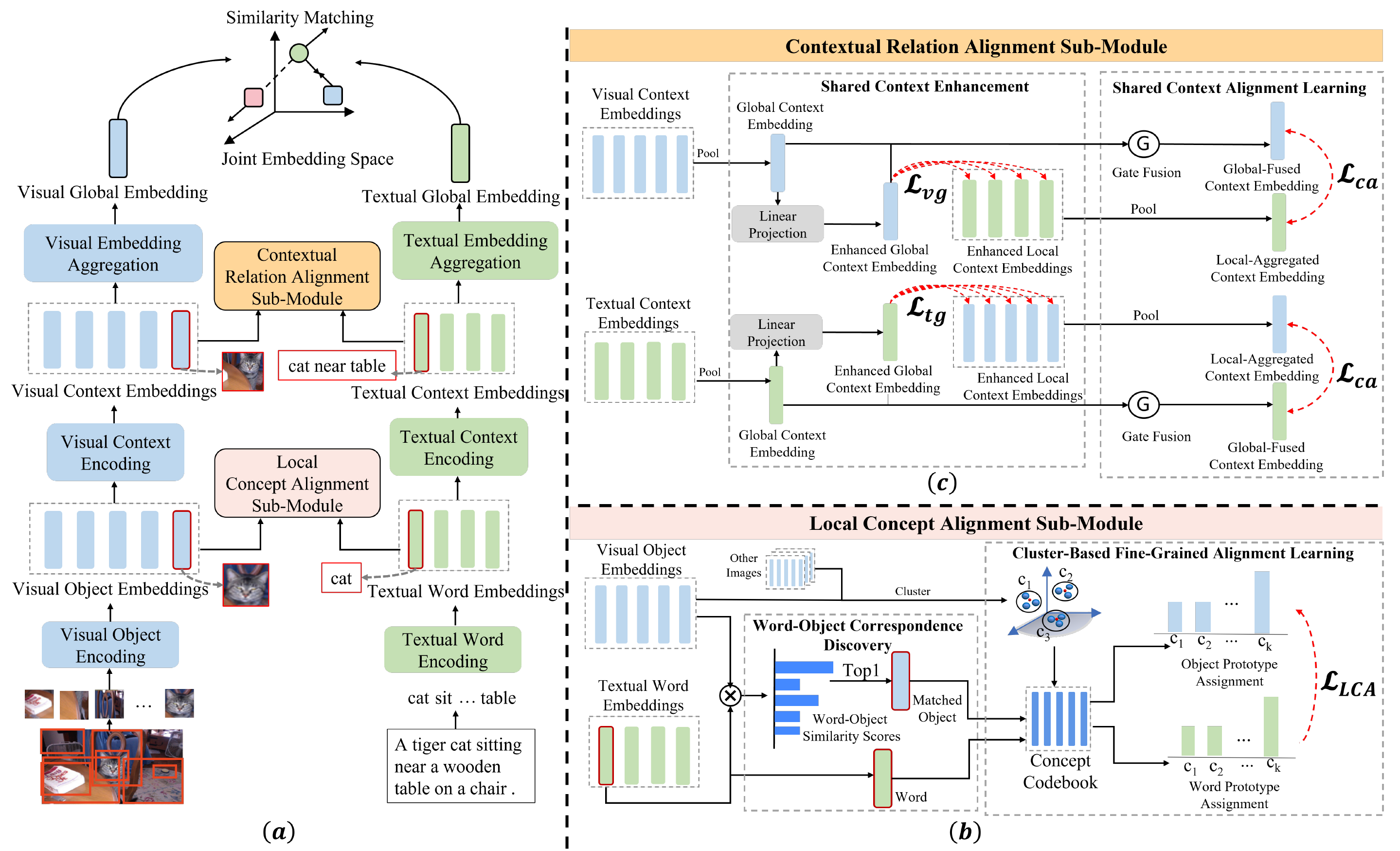}  
\caption{The overview of SelfAlign applied to independent-embedding models. SelfAlign consists of two sub-modules: Local Concept Alignment sub-module and Contextual Relation Alignment sub-module.  (a) illustrates the hierarchical encoding stage of independent-embedding models and the inject position for the two sub-modules of SelfAlign. (b) describes the Local Concept Alignment sub-module and (c) describes the Contextual Relation Alignment sub-module.} 
\label{fig_method}
\end{figure*}

We propose a module SelfAlign to explore the multi-grained correspondences in the different layers of independent-embedding models. \textcolor{black}{Independent-embedding models mainly consist of a visual encoder and a textual encoder, as shown in Figure \ref{Fig:intro} (a).} \textcolor{black}{Though the encoders are various from adopting different encoding architectures such as GCN \cite{kipf2016semi} and self-attention \cite{wu2019learning,qu2020context}, the encoding process for each modality typically includes three stages as shown in the Figure \ref{fig_method} (a), visual object encoding and textual word encoding stage, visual and textual context encoding stage and visual and textual embedding aggregation stage.}
\textcolor{black}{However, independent-embedding models only conduct global \textcolor{black}{image-text} alignment at the embedding aggregation stage and overlook fine-grained alignment in the first two encoding stages.} \textcolor{black}{To improve their retrieval accuracy, we design two sub-modules in SelfAlign injected in the first two encoding stages respectively: 1) \textbf{L}ocal \textbf{C}oncept \textbf{A}lignment (\textbf{LCA}) sub-module for local conceptual level alignment between visual objects and textual words, 2) \textbf{C}ontextual \textbf{R}elation \textbf{A}lignment (\textbf{CRA}) sub-module for contextual level alignment.}

In this section, we first describe \textcolor{black}{the single modal embedding extraction} approaches in the independent-embedding models in Section \ref{sec_fea_extra}. \textcolor{black}{In Section \ref{sec_lca}, we introduce the LCA sub-module, which learns the concept-level word-object correspondences at the object and word encoding stage. We then introduce the CRA sub-module in Section \ref{sec_cra} to explore context-level alignment in the context encoding stage.} Since SelfAlign is model-agnostic and applicable to independent-embedding models, we case study on two representative baseline models, VSRN \cite{li2019visual} and CAMERA \cite{qu2021dynamic}, which is introduced in Section \ref{sec_case_study} and Section \ref{sec_objective}, respectively.

\subsection{\textcolor{black}{Single-Modal Embedding Extraction}}
\label{sec_fea_extra}
\textbf{Image Embedding Extraction.} For each input image $I$, 
recent works \cite{lee2018stacked,li2019visual} usually employ an off-the-shelf object detection model, such as Faster R-CNN \cite{ren2016faster}, to detect $M$ objects
$O=\left\{o_{i}\right\}^{M}_{i=1}$, where each object $o_{i}$ is represented by an object feature embedding $\bm{o}_{i} \in \mathbb{R}^{d_{o}}$. Then a linear projection is utilized to transform $\bm{o}_{i}$ into a $h$-dimensional embedding. Then the embedding of the image is represented by a set of object embeddings $\bm{V}^{l}=\left\{ \bm{v}_{i}^{l}\right\}^{M}_{i=1}$. We name this encoding process as visual object encoding stage. Then different works utilize various approaches such as GCN \cite{li2019visual} or Transformers \cite{li2019visual} to model the relationships between objects and obtain contextualized object embeddings, named as visual context encoding stage. Here, we omit the computation details and simplify them as the visual context encoder, and the output of the context encoder is denoted as visual context embeddings $\bm{V}^{c}=\left\{ \bm{v}^{c}_{i}\right\}^{M}_{i=1}$. Finally, the visual global embedding $\bm{V}^{g}$ is obtained by integrating the context embeddings $\bm{V}^{c}$, which is named as visual embedding aggregation stage. 

\textbf{Text Embedding Extraction.} For each input text $T$, the word-level embeddings $\bm{W}=\left\{ \bm{w}_{i}\right\}^{D}_{i=1}$ is generally obtanined from the pre-trained word embeddings \cite{li2019visual, lee2018stacked}, such as BERT \cite{devlin2018bert}. 
Then an embedding layer is also employed to transform $\bm{w}_{i}$ to a $h$-dimensional vector. The word embedding set is denoted as $\bm{T}^{l}=\left\{ \bm{t}_{i}^{l}\right\}^{D}_{i=1}$. We name this encoding process as the textual word encoding stage. 
Then different works utilize different approaches, such as GRU \cite{lee2018stacked, li2019visual} and Transformer \cite{qu2020context, ALIGN}, to capture the contextualized word embeddings, named as the textual context encoding stage. We denote the output of this stage as textual context embeddings $\bm{T}^{c}=\left\{ \bm{t}^{c}_{i}\right\}^{D}_{i=1}$, \textcolor{black}{where $D$ denotes the number of the words in the sentence.} The final global embedding $\bm{T}^{g}$ is obtained by aggregating the context embeddings $\bm{T}^{c}$, named as texual embedding aggregation stage.

\subsection{Local Concept Alignment Sub-Module}
\label{sec_lca}
Local Concept Alignment (LCA) sub-module is injected into the object and word encoding layer of the baseline model and aims to force the consistency between the visual and textual concept embeddings, as presented in Figure \ref{fig_method} (b). \textcolor{black}{We regard the object regions as the visual concepts and the whole words in the sentences as textual concepts to make full use of information in the whole sentence.}
Since there are no word-object pair annotations, a word-object correspondence discovery strategy is first proposed to build pseudo word-object correspondences in LCA. Then a cluster-based fine-grained alignment is designed to force the consistency between the selected matched word-object pair by comparing their cluster assignments. Though there are no annotations \textcolor{black}{of} the word-object pairs, our concept alignment is under the constraint of global image-text matching supervision. 

\textbf{Word-Object Correspondence Discovery.} We take the visual object embeddings $\bm{V}^{l}$ and word embeddings $\bm{T}^{l}$ as the input of the LCA sub-module. Since there are no word-object annotations, we first estimate the word-object correspondences from image-text pairs. Specifically, we compute the cosine similarities between the words and objects and choose the most similar object as the correspondence for each word. 
\textcolor{black}{Notably, since the texts of an image is the linguistic expression of the image itself \cite{young2014image, lin2014microsoft}, each word \textcolor{black}{is} able to find at least one matched object region given an image-text pair. We cannot guarantee each object in the image has corresponding words in the text. Therefore, we align each word to an object rather than the inverse direction.}

Formally, given the $i$-th word embedding $\bm{t}_{i}^{l}$, we compute its cosine similarities with the region embeddings $\bm{V}^{l}$ and apply the argmax operation to select the most matched object region:
\begin{equation}
\setlength{\abovedisplayskip}{5pt}
\setlength{\belowdisplayskip}{5pt}
\label{loc_argmax}
 \boldsymbol{v}_{j^{+}}^{l}=\underset{j \in\left[1, M\right]}{\arg \max } \ {\rm cos}\left( \bm{t}_{i}^{l}, \bm{v}_{j}^{l}\right)   
\end{equation}
where $\bm{v}_{j^{+}}$ is the selected matched image object region for $\bm{t}_{i}$ and ${\rm cos}(\cdot,\cdot)$ represents the cosine similarity function. We utilize the whole word-object pairs obtained by all words in a batch as pseudo-supervised alignment information for local concept-level alignment. 

\textbf{Cluster-based Fine-grained Alignment.} 
\textcolor{black}{Cluster-based fine-grained alignment aims to enforce the consistency between each word embedding and the matched object embedding selected by the word-object correspondence discovery strategy. Inspired by the recent success of self-supervised learning \cite{simclr, DBLP:conf/icml/TianCG21}, we treat the selected matched word-object pair as two different views depicting the same conceptual semantic and utilize cluster-based contrastive learning \cite{caron2020unsupervised, DBLP:conf/eccv/CaronBJD18} to align their embeddings across different modalities. The basic idea lies in that we cluster object region embeddings to form a concept dictionary, where each cluster center represents a concept. The embeddings of the cluster centers are more stable and representative to represent concepts compared with the object region embeddings in each image. Each region embedding and word embedding are then mapped to the concept dictionary and represented by concept assignments. In this way, we represent regions and words at the semantic-consistent concept level instead of the expression-variant instance level. After that, we utilize cross-prediction to align the region and word embeddings. Notably, there is an alternative solution to align the embeddings, named as individual-based contrastive learning \cite{simclr,moco}. However, individual-based contrastive learning requires a large number of paired negative-positive samples, and it is challenging to select the appropriate negative samples without ground-truth word-object annotations. Thus, to efficiently and effectively align the word-object embeddings, we adopt cluster-based contrastive learning without requiring numerous computations. It is an alternative to contrasting multiple views by comparing their cluster assignments instead of their instance-level embeddings \cite{caron2020unsupervised}.}

\textcolor{black}{The cluster-based fine-grained alignment process contains two main steps: (1) We first perform \textit{concept codebook construction} to obtain the concept assignments, where the concept codebook is a collection of learnable prototypes based on object region embeddings. (2) Then \textit{cross-prediction learning} is conducted from word concept assignments to object concept assignments to align the word-object embeddings. 
}

\textcolor{black}{\textit{Step 1: concept codebook construction.}} To learn the concept codebook and let the learning process of cluster centers be synchronous with the whole network, we use the online clustering method as in \cite{caron2020unsupervised}. Following \cite{caron2020unsupervised}, we define the trainable concept codebook $\bm{C}=\left\{ \bm{c}_{i}\right\}^{K}_{i=1}$, 
where $K$ represents the number of concept centers and each center $\bm{c}_{i}$ is a $h$-dimensional vector as visual and textual concept embeddings.

\textcolor{black}{\textit{Step 2: cross-prediction learning.} We use cross-entropy loss between object and word concept assignments to set up this cross-prediction problem. Firstly, given the concept codebook,} the visual concept assignment $\bm{v}_{i}^{P} \in \mathbb{R}^{K}$ for the object embedding $\bm{v}_{i}^{l}$ can be obtained by mapping $\bm{v}_{i}^{l}$ to the concept codebook. Each scalar $\bm{p}_{i}^{k}$ in $\bm{v}_{i}^{P}$ represents the probability that $\bm{v}_{i}$ belongs to the $k$-th prototype $\bm{c}_{k}$ and is obtained by taking a softmax of the cosine similarity of $\bm{v}_{i}^{l}$ and $\bm{c}_{k}$ as follows:
\begin{equation}
\setlength{\abovedisplayskip}{5pt}
\setlength{\belowdisplayskip}{5pt}
\label{code_assignment}
\bm{p}_{i}^{k}=\frac{\exp \left({\rm cos}\left(\bm{v}_{i}^{l}, \bm{c}_{k}\right) / \tau_{1}\right)}{\sum_{k=1}^{K} \exp  \left({\rm cos}\left(\bm{v}_{i}^{l}, \bm{c}_{k}\right) / \tau_{1}\right)}
\end{equation}
where $\tau_{1}$ is a temperature parameter. 
Similarly, the word assignment \textcolor{black}{is} computed by mapping each word embedding $\bm{t}_{i}^{l}$ to the concept codebook, and each scalar of the probability that $\bm{t}_{i}^{l}$ belongs to the $k$-th prototype, denoted as $\bm{q}_{i}^{k}$.

Then cross-prediction from word concept assignments to the corresponding object concept assignments is performed to align the selected word-object pair. The prediction problem is optimized by the cross-entropy loss as follows:
\begin{equation}
\setlength{\abovedisplayskip}{5pt}
\setlength{\belowdisplayskip}{5pt}
\label{loc_q}
\begin{aligned}
&\mathcal{L}_{i}=-\sum_{k=1}^{K} \bm{p}_{j^{+}}^{k} \log  \bm{q}_{i}^{k}
\end{aligned}
\end{equation}
where $\bm{p}_{j^{+}}^{k}$ is the object concept assignment of the matched visual object embedding $\bm{v}_{i^{+}}^{l}$. Finally, we aggregate the loss values in Equation \ref{loc_q} over all the words $D$ and result in the following loss function to optimize the feature encoders:
\begin{equation}
\label{loc_loss_sum}
\begin{aligned}
&\mathcal{L}_{LCA}=\frac{1}{D} {\sum_{i=1}^{D}\mathcal{L}_{i}}\\
\end{aligned}
\end{equation}

\subsection{Contextual Relation Alignment Sub-Module}
\label{sec_cra}
Since semantically similar concepts have different semantics in different contexts, Context Relation Alignment (CRA) sub-module is further proposed to capture context-level semantic correspondences in the context embedding layer, as presented in Figure \ref{fig_method} (c). CRA first performs shared context enhancement to capture the shared context-level information and suppress the irrelevant information between the images and texts. Shared context alignment is then conducted to achieve contextual-level alignment. 
\textbf{Shared Context Enhancement.} Given the visual context embeddings $\bm{V}^{c}=\left\{ \bm{v}^{c}_{i}\right\}^{M}_{i=1}$ and text context embeddings $\bm{T}^{c}=\left\{ \bm{t}^{c}_{i}\right\}^{D}_{i=1}$, we acquire the global context embedding $\bm{v}^{c}_{g}$ and $\bm{t}^{c}_{g}$ for contextual relation alignment by adopting the average-pooling over $\bm{V}^{c}$ and $\bm{T}^{c}$ respectively. Here, we design two symmetric contrastive mechanism to learn the shared context information from visual perspective and textual perspective: contrasting between the visual global context embedding and the textual local context embeddings, denoted as V-global/T-local and contrasting between the textual global context embedding and the visual local context embeddings, denoted as T-global/V-local. Under the global supervision from one modality, the relevant local semantics of objects and relationships in the other modality will be strengthened while irrelevant semantics will be weakened.

Formally, a fully connected layer is first utilized to map the global context embedding into the corresponding local context space as the global supervision: 
\begin{equation}
\label{c_fc}
\begin{gathered}
\bm{t}_{s}^{c}=\sigma\left({\rm BN}\left(\mathbf{W}_{1}\bm{t}_{g}^{c}+{b_{1}}\right)\right) \\
\bm{v}_{s}^{c}=\sigma\left({\rm BN}\left(\mathbf{W}_{2}\bm{v}_{g}^{c}+{b_{2}}\right)\right)
\end{gathered}
\end{equation}
where $\sigma$ is the ReLU activation and BN is the batch normalization. $\mathbf{W}_{1}$, $\mathbf{W}_{2}$, $b_{1}$ and $b_{2}$ are learnable parameters. Meanwhile, the batch normalization and ReLU activation operations are employed on the local contextual embeddings ${\bm{v}_{i}^{c}}, {\bm{t}_{i}^{c}}$, and they are then denoted as ${\bm{v}_{i}^{*}}, {\bm{t}_{i}^{*}}$ in contrastive learning.

Given the global context and local context embeddings, the T-global/V-local contrastive learning and V-global/T-local contrastive learning is conducted in parallel. For T-global/V-local contrastive learning, we consider the textual global supervision embedding $\bm{t}_{s}^{c}$ and the visual local contextual embeddings $\bm{v}_{i}^{*}$ that come from an image-text pair as a positive sample, while the embeddings from unpaired image-text pairs as negative samples. Then we define the T-global/V-local contrastive loss as:
\begin{equation}
\label{ctx_tgvl}
\mathcal{L}_{tg}=-\frac{1}{M} \sum_{i=1}^{M} \log \frac{\exp \left({\rm cos}\left(\bm{t}_{s}^{c}, \bm{v}_{i}^{*}\right) / \tau_{2}\right)}{\sum_{j=1}^{N} \exp \left({\rm cos}\left(\bm{t}_{s}^{c}, \tilde{\bm{v}}_{j}^{*}\right) / \tau_{2}\right)}
\end{equation}
where $\tau_{2}$ is a temperature hyper-parameter and $M$ is the number of visual regions. $\left\{ \tilde{\bm{v}}_{j}^{*}\right\}^{N}_{j=1}$ is a set of negative visual local embeddings. $N$ is the number of negatives. Similarly, the V-global/T-local contrastive loss is defined as:
\begin{equation}
\label{ctx_vgtl}
\mathcal{L}_{vg}=-\frac{1}{D} \sum_{i=1}^{D} \log \frac{\exp \left({\rm cos}\left(\bm{v}_{s}^{c}, \bm{t}_{i}^{*}\right) / \tau_{2}\right)}{\sum_{j=1}^{N} \exp \left({\rm cos}\left(\bm{v}_{s}^{c}, \tilde{\bm{t}}_{j}^{*}\right) / \tau_{2}\right)}
\end{equation}
where $\left\{ \tilde{\bm{t}}_{j}^{*}\right\}^{N}_{j=1} $ is negative textually local context embeddings. $D$ is the number of textual words. \textcolor{black}{We select the $N$ negative local embeddings that are most similar to the global embedding of the other modality ranked by cosine similarity. As a result, these samples are hard negative samples which are beneficial for learning high-quality representations of the anchor sample \cite{high-quality}.} The loss for shared context enhancement is defined as:
\begin{equation}
\label{loss_cs}
 \mathcal{L}_{cs}=\frac{1}{2}\left(\mathcal{L}_{tg}+\mathcal{L}_{vg}\right)   
\end{equation}

\textbf{Shared Context Alignment.} 
Shared context alignment aims to perform contextual level alignment based on the enhanced global context embedding and local context embeddings. Specifically, both of the V-global/T-local and the T-global/V-local contrastive learning enhance the global context embeddings $\bm{v}^{c}_{g}$ and $\bm{t}^{c}_{g}$ from local and global perspectives and introduce a paired final global context embeddings for context alignment. Taking T-global/V-local contrastive learning as an example, it enables the encoder to filter out irrelevant local contextual information which is not aligned with global contextual information of the other modality. We further use average-pooling on visual enhanced local context embeddings $\left\{ \bm{v}_{i}^{*} \right\}_{i=1}^{M}$ to form the final visual context embedding, denoted as the visual local-aggregated context embedding $\bm{v}_{g}^{*}$. Given the text global context embedding $\bm{t}_{g}^{c}$ and its enhanced global context embedding $\bm{t}_{s}^{c}$, we perform the fusion operation as follows:
\begin{equation}
\label{fusion}
\begin{aligned}
&\bm{t}_{f}^{c}=g \cdot \bm{t}_{s}^{c}+(1-g) \cdot \bm{t}_{g}^{c} \\
&g=\operatorname{sigmoid}\left(\mathbf{W}_{g} \left[\bm{t}_{s}^{c}, \bm{t}_{g}^{c}\right]+\rm b_{g}\right)
\end{aligned}
\end{equation}
where $\bm{t}_{f}^{c}$ denotes the textual global-fused context embedding obtained by the text global context information $\bm{t}_{g}^{c}$ and its enhanced global context information $\bm{t}_{s}^{c}$ from visual modality. $g$ is a gating value to adaptively balance the importance of $\bm{t}_{g}^{c}$ and $\bm{t}_{s}^{c}$. $[\cdot, \cdot]$ means the concatenation operation.

Similarly, as for V-global/T-local contrastive learning, we obtain the textual local-aggregated context embedding $\bm{t}_{g}^{*}$ based on $\left\{ \bm{t}_{i}^{*} \right\}_{i=1}^{D}$. By fusing $\bm{v}_{g}^{c}$ and $\bm{v}_{s}^{c}$, we get the visual global-fused context embedding $\bm{v}_{f}^{c}$ similar to Equation \ref{fusion}. 

Based on these final global context embeddings, we obtain the contextual-level matching score for a given image-text pair ($I$, $T$), which is defined as:
\begin{equation}
\label{ctx_sim}
S_{c}(I,T)=\frac{1}{2}\left({\rm cos}(\bm{t}^{c}_{f}, \bm{v}^{*}_{g}) + {\rm cos}(\bm{v}^{c}_{f},\bm{t}^{*}_{g})\right)
\end{equation}

Then we use the hinge-based triplet ranking loss \cite{faghri2017vse++} to enforce the contextual similarity of matched image-text pair to be higher than unmatched ones for contextual alignment: 
\begin{equation}
\label{loss_ca}
\begin{aligned}
\mathcal{L}_{ca}=max\left(0, \alpha+S_{c}\left(\mathit{I}, \mathrm{\widetilde{\mathit{T}}}\right)-S_{c}\left(\mathrm{\mathit{I}}, \mathrm{\mathit{T}}\right)\right) \\
+max\left(0,\alpha+S_{c}\left(\mathrm{\widetilde{\mathit{I}}}, \mathrm{\mathit{T}}\right)-S_{c}\left(\mathrm{\mathit{I}}, \mathrm{\mathit{T}}\right)\right) \\
\end{aligned}
\end{equation}
where $\widetilde{T}=\operatorname{argmax}_{d \neq T} s(I, d)$ and $\widetilde{I}=\operatorname{argmax}_{j \neq I} s(j, T)$ are the hardest negatives in a mini-batch for a positive pair ($I$, $T$), and $\alpha$ is the margin parameter. Taking the loss objective for shared context enhancement in Equation \ref{loss_cs} together, we define the total loss objective in CRA sub-module as:
\begin{equation}
    \label{loss_cra}
  \mathcal{L}_{CRA}=\mathcal{L}_{cs}+ \mathcal{L}_{ca}
\end{equation}

\subsection{SelfAlign with Typical Independent-Embedding Models}
\label{sec_case_study}
To prove the effectiveness of our module SelfAlign, we case study on two typical independent-embedding models, the widely compared model VSRN \cite{li2019visual} and the state-of-the-art independent-embedding model CAMERA \cite{qu2020context}.  

\textbf{VSRN with SelfAlign.} 
For image encoding, VSRN \cite{li2019visual} uses the object embeddings from an object detection model as visual inputs, followed by an FC layer and a graph convolution network with four layers to perform object-relation reasoning. Finally, VSRN utilizes a GRU unit to \textcolor{black}{obtain} the visual global embedding. For text encoding, VSRN exploits a word embedding layer to encode word-level embeddings and adopts an LSTM to obtain the contextualized word embeddings. Finally, VSRN utilizes the final hidden state of the LSTM as the textual global embedding. For the loss function, VSRN utilizes the sum of a matching loss on global embeddings via triplet ranking loss and a generation objective from images to texts to jointly align the images and texts. Here, we denote the loss function of VSRN as $\mathcal{L}_{base}$. We take the outputs of \textcolor{black}{the FC layer in the image encoder and the outputs of word-embedding layer in the text encoder} into LCA sub-module. And we regard the outputs of the last GCN layer and the LSTM as the input of the CRA sub-module.  

\textbf{CAMERA with SelfAlign.}
For the image encoder, CAMERA concatenates the object embeddings and the position embeddings of each object as visual inputs and an FC layer is followed to obtain the visual object embeddings. The word embedding inputs are obtained from the output of pre-trained BERT \cite{devlin2018bert}. 
Then CAMERA adopts a self-attention layer to perform relation reasoning for image and text respectively. For the loss function, CAMERA adopts the sum of a triplet ranking loss on global embeddings and a diversity regularization loss for cross-modal alignment and the summarization of multi-view descriptions. Here, we also denote the loss function of CAMERA as $\mathcal{L}_{base}$. 
And we take the outputs of the FC layer as LCA sub-module inputs, taking the outputs of the self-attention layer as the inputs of the CRA sub-module. 

\subsection{Model Training and Inference}
\label{sec_objective}
The final training loss for baseline models with SelfAlign is defined as:
 \begin{equation}
\label{all_loss}
  \mathcal{L}=\mathcal{L}_{base}+\mathcal{L}_{LCA}+\mathcal{L}_{CRA}
\end{equation}
where $\mathcal{L}_{base}$ means the loss objectives of the baseline model. $\mathcal{L}_{LCA}$ and $\mathcal{L}_{CRA}$ are defined in Equation \ref{loc_loss_sum} and Equation \ref{loss_cra}.

In the inference stage, the similarity for each image-text pair of baseline with SelfAlign is computed as follows:
\begin{equation}
\label{infer_sim}
    S(I,T)= S_{b}(I,T) + S_{c}(I,T) 
\end{equation}
where $S_{b}(I,T)$ is the similarity score evaluated by the baseline model, and $S_{c}(I,T)$ denotes our proposed contextual-level similarity score according to Equation \ref{ctx_sim}. It is noteworthy that since our module keeps the advantage of the independent encoding framework of the independent-embedding models, the queried image or text embeddings can pre-computed offline before the inference stage.

\section{Experiment}
\textbf{Datasets.} We conduct extensive experiments on two benchmark datasets in image-text retrieval: Flickr30K \cite{young2014image} and MS-COCO \cite{lin2014microsoft}. Flickr30K consists of 31,783 images collected from the Flickr website and each image is associated with 5 sentences. Following the settings in \cite{faghri2017vse++, lee2018stacked}, we utilized 1,000 images for validation, 1,000 images for testing, and the rest for training. MS-COCO contains 123,287 images with 5 captions for each image. Following \cite{faghri2017vse++, lee2018stacked}, we take 113,287 images for training, 5,000 images for validation, and 5,000 images for testing, and the results are reported by both averaging over 5 folds of 1K test set images and testing on the full 5K test images as in \cite{faghri2017vse++, lee2018stacked}.

\textbf{Evaluation Metrics.} To compare with the state-of-the-art models, we adopt the commonly used evaluation metrics in all datasets as \cite{faghri2017vse++,lee2018stacked,li2019visual}. Namely, we adopt Recall at K denoted as R@K to evaluate the performance on both text retrieval (retrieve the most related text given an image query) and image retrieval (retrieve the most related image given a text query) tasks. R@K means the percentage of queries that are correctly matched in the top-\textit{K} ranking list. We report R@1, R@5, R@10 for all datasets as in \cite{li2019visual}. Besides, to comprehensively reveal the overall retrieval performance, we also report another metric R@sum as in \cite{chen2020imram}, defined as the summation of all R@K values in both retrieval tasks. 

\textbf{Implementation Details.} To perform a fair comparison, for VSRN and CAMERA, we completely preserve their network structures and model settings such as training batch size and other model-related hyper-parameter settings as stated in their original work. We only inject our module into the two baselines as introduced in Section \ref{sec_case_study}. For both baseline models, the softmax temperature $\tau_{1}$ in the LCA sub-module is set to 0.1 as in \cite{caron2020unsupervised}. The number of concept classes $K$ is set to 1024, and the number of negative samples \textit{N} and the softmax temperature $\tau_{2}$ in CRA sub-module are set to 512 and 0.7. All of our experiments are conducted on a single NVIDIA Tesla GPU with 24GB memory and implemented in PyTorch.
\subsection{State-of-the-Art Comparison}
\label{sec_sota}
\begin{table*}[]
\caption{Comparison with existing independent-embedding models and interactive-embedding models. Our re-implemented independent-embedding models are denoted by the superscript `*'. The highest retrieval accuracy in each block is marked with \uline{underline}. The accuracy of our models is marked with \textcolor{black}{blue} color when is better than the baseline model.}
\renewcommand\arraystretch{1.2}
\resizebox{\textwidth}{!}{%
\begin{tabular}{lccccccccccccccccccccc}
\hline
\multicolumn{1}{l|}{} &
  \multicolumn{7}{c|}{\textbf{Flickr30k}} &
  \multicolumn{7}{c|}{\textbf{MS-COCO 1K}} &
  \multicolumn{7}{c}{\textbf{MS-COCO 5K}} \\ \cline{2-22} 
\multicolumn{1}{l|}{} &
  \multicolumn{3}{c|}{\textbf{Text Retrieval}} &
  \multicolumn{3}{c|}{\textbf{Image Retrieval}} &
  \multicolumn{1}{c|}{\cellcolor[HTML]{C0C0C0}} &
  \multicolumn{3}{c|}{\textbf{Text Retrieval}} &
  \multicolumn{3}{c|}{\textbf{Image Retrieval}} &
  \multicolumn{1}{c|}{\cellcolor[HTML]{C0C0C0}} &
  \multicolumn{3}{c|}{\textbf{Text Retrieval}} &
  \multicolumn{3}{c|}{\textbf{Image Retrieval}} &
  \cellcolor[HTML]{C0C0C0} \\ \cline{2-7} \cline{9-14} \cline{16-21}
\multicolumn{1}{l|}{\multirow{-3}{*}{\textbf{Model}}} &
  \multicolumn{1}{c|}{\textbf{R@1}} &
  \multicolumn{1}{c|}{\textbf{R@5}} &
  \multicolumn{1}{c|}{\textbf{R@10}} &
  \multicolumn{1}{c|}{\textbf{R@1}} &
  \multicolumn{1}{c|}{\textbf{R@5}} &
  \multicolumn{1}{c|}{\textbf{R@10}} &
  \multicolumn{1}{c|}{\multirow{-2}{*}{\cellcolor[HTML]{C0C0C0}\textbf{R@sum}}} &
  \multicolumn{1}{c|}{\textbf{R@1}} &
  \multicolumn{1}{c|}{\textbf{R@5}} &
  \multicolumn{1}{c|}{\textbf{R@10}} &
  \multicolumn{1}{c|}{\textbf{R@1}} &
  \multicolumn{1}{c|}{\textbf{R@5}} &
  \multicolumn{1}{c|}{\textbf{R@10}} &
  \multicolumn{1}{c|}{\multirow{-2}{*}{\cellcolor[HTML]{C0C0C0}\textbf{R@sum}}} &
  \multicolumn{1}{c|}{\textbf{R@1}} &
  \multicolumn{1}{c|}{\textbf{R@5}} &
  \multicolumn{1}{c|}{\textbf{R@10}} &
  \multicolumn{1}{c|}{\textbf{R@1}} &
  \multicolumn{1}{c|}{\textbf{R@5}} &
  \multicolumn{1}{c|}{\textbf{R@10}} &
  \multirow{-2}{*}{\cellcolor[HTML]{C0C0C0}\textbf{R@sum}} \\ \hline
\multicolumn{22}{c}{\textbf{Independent-embedding Models}} \\ \hline
\multicolumn{1}{l|}{Order \cite{vendrov2015order} (2016)} &
  \multicolumn{1}{c|}{-} &
  \multicolumn{1}{c|}{-} &
  \multicolumn{1}{c|}{-} &
  \multicolumn{1}{c|}{-} &
  \multicolumn{1}{c|}{-} &
  \multicolumn{1}{c|}{-} &
  \multicolumn{1}{c|}{\cellcolor[HTML]{C0C0C0}-} &
  \multicolumn{1}{c|}{46.7} &
  \multicolumn{1}{c|}{78.6} &
  \multicolumn{1}{c|}{88.9} &
  \multicolumn{1}{c|}{37.9} &
  \multicolumn{1}{c|}{73.7} &
  \multicolumn{1}{c|}{85.9} &
  \multicolumn{1}{c|}{\cellcolor[HTML]{C0C0C0}411.7} &
  \multicolumn{1}{c|}{-} &
  \multicolumn{1}{c|}{-} &
  \multicolumn{1}{c|}{-} &
  \multicolumn{1}{c|}{-} &
  \multicolumn{1}{c|}{-} &
  \multicolumn{1}{c|}{-} &
  \cellcolor[HTML]{C0C0C0}- \\
\multicolumn{1}{l|}{2WayNet \cite{eisenschtat2017linking} (2017)} &
  \multicolumn{1}{c|}{49.8} &
  \multicolumn{1}{c|}{67.5} &
  \multicolumn{1}{c|}{-} &
  \multicolumn{1}{c|}{36.0} &
  \multicolumn{1}{c|}{55.6} &
  \multicolumn{1}{c|}{-} &
  \multicolumn{1}{c|}{\cellcolor[HTML]{C0C0C0}208.9} &
  \multicolumn{1}{c|}{55.8} &
  \multicolumn{1}{c|}{75.2} &
  \multicolumn{1}{c|}{0.0} &
  \multicolumn{1}{c|}{39.7} &
  \multicolumn{1}{c|}{63.3} &
  \multicolumn{1}{c|}{0.0} &
  \multicolumn{1}{c|}{\cellcolor[HTML]{C0C0C0}234.0} &
  \multicolumn{1}{c|}{23.3} &
  \multicolumn{1}{c|}{50.5} &
  \multicolumn{1}{c|}{65.0} &
  \multicolumn{1}{c|}{18.0} &
  \multicolumn{1}{c|}{43.6} &
  \multicolumn{1}{c|}{57.6} &
  \cellcolor[HTML]{C0C0C0}258.0 \\
\multicolumn{1}{l|}{VSE++ \cite{faghri2017vse++} (2018)} &
  \multicolumn{1}{c|}{52.9} &
  \multicolumn{1}{c|}{79.1} &
  \multicolumn{1}{c|}{87.2} &
  \multicolumn{1}{c|}{39.6} &
  \multicolumn{1}{c|}{69.6} &
  \multicolumn{1}{c|}{79.5} &
  \multicolumn{1}{c|}{\cellcolor[HTML]{C0C0C0}407.9} &
  \multicolumn{1}{c|}{64.6} &
  \multicolumn{1}{c|}{89.1} &
  \multicolumn{1}{c|}{95.7} &
  \multicolumn{1}{c|}{52.0} &
  \multicolumn{1}{c|}{83.1} &
  \multicolumn{1}{c|}{92.0} &
  \multicolumn{1}{c|}{\cellcolor[HTML]{C0C0C0}476.5} &
  \multicolumn{1}{c|}{41.3} &
  \multicolumn{1}{c|}{69.2} &
  \multicolumn{1}{c|}{81.2} &
  \multicolumn{1}{c|}{30.3} &
  \multicolumn{1}{c|}{59.1} &
  \multicolumn{1}{c|}{72.4} &
  \cellcolor[HTML]{C0C0C0}353.5 \\
\multicolumn{1}{l|}{GXN \cite{wang2017adversarial} (2018)} &
  \multicolumn{1}{c|}{56.8} &
  \multicolumn{1}{c|}{-} &
  \multicolumn{1}{c|}{89.6} &
  \multicolumn{1}{c|}{41.5} &
  \multicolumn{1}{c|}{-} &
  \multicolumn{1}{c|}{80.1} &
  \multicolumn{1}{c|}{\cellcolor[HTML]{C0C0C0}-} &
  \multicolumn{1}{c|}{68.5} &
  \multicolumn{1}{c|}{-} &
  \multicolumn{1}{c|}{97.9} &
  \multicolumn{1}{c|}{56.6} &
  \multicolumn{1}{c|}{-} &
  \multicolumn{1}{c|}{94.5} &
  \multicolumn{1}{c|}{\cellcolor[HTML]{C0C0C0}-} &
  \multicolumn{1}{c|}{42.0} &
  \multicolumn{1}{c|}{-} &
  \multicolumn{1}{c|}{84.7} &
  \multicolumn{1}{c|}{31.7} &
  \multicolumn{1}{c|}{-} &
  \multicolumn{1}{c|}{74.6} &
  \cellcolor[HTML]{C0C0C0}- \\
\multicolumn{1}{l|}{VSRN \cite{li2019visual} (2019)} &
  \multicolumn{1}{c|}{70.4} &
  \multicolumn{1}{c|}{89.2} &
  \multicolumn{1}{c|}{93.7} &
  \multicolumn{1}{c|}{53.0} &
  \multicolumn{1}{c|}{77.9} &
  \multicolumn{1}{c|}{85.7} &
  \multicolumn{1}{c|}{\cellcolor[HTML]{C0C0C0}469.9} &
  \multicolumn{1}{c|}{74.0} &
  \multicolumn{1}{c|}{94.3} &
  \multicolumn{1}{c|}{97.8} &
  \multicolumn{1}{c|}{60.8} &
  \multicolumn{1}{c|}{88.4} &
  \multicolumn{1}{c|}{94.1} &
  \multicolumn{1}{c|}{\cellcolor[HTML]{C0C0C0}509.4} &
  \multicolumn{1}{c|}{50.3} &
  \multicolumn{1}{c|}{79.6} &
  \multicolumn{1}{c|}{87.9} &
  \multicolumn{1}{c|}{37.9} &
  \multicolumn{1}{c|}{68.5} &
  \multicolumn{1}{c|}{79.4} &
  \cellcolor[HTML]{C0C0C0}403.6 \\
\multicolumn{1}{l|}{CAMERA \cite{qu2020context} (2020)} &
  \multicolumn{1}{c|}{76.5} &
  \multicolumn{1}{c|}{{\ul 95.1}} &
  \multicolumn{1}{c|}{97.2} &
  \multicolumn{1}{c|}{58.9} &
  \multicolumn{1}{c|}{84.7} &
  \multicolumn{1}{c|}{90.2} &
  \multicolumn{1}{c|}{\cellcolor[HTML]{C0C0C0}502.6} &
  \multicolumn{1}{c|}{75.9} &
  \multicolumn{1}{c|}{95.5} &
  \multicolumn{1}{c|}{98.6} &
  \multicolumn{1}{c|}{62.3} &
  \multicolumn{1}{c|}{{\ul 90.9}} &
  \multicolumn{1}{c|}{{\ul 95.8}} &
  \multicolumn{1}{c|}{\cellcolor[HTML]{C0C0C0}519.0} &
  \multicolumn{1}{c|}{53.1} &
  \multicolumn{1}{c|}{81.3} &
  \multicolumn{1}{c|}{89.8} &
  \multicolumn{1}{c|}{39.0} &
  \multicolumn{1}{c|}{70.5} &
  \multicolumn{1}{c|}{81.5} &
  \cellcolor[HTML]{C0C0C0}415.2 \\ \hline
\multicolumn{22}{c}{\textbf{Interactive-embedding Models}} \\ \hline
\multicolumn{1}{l|}{SCAN\_ensemble \cite{lee2018stacked} (2018)} &
  \multicolumn{1}{c|}{67.4} &
  \multicolumn{1}{c|}{90.3} &
  \multicolumn{1}{c|}{95.8} &
  \multicolumn{1}{c|}{48.6} &
  \multicolumn{1}{c|}{77.7} &
  \multicolumn{1}{c|}{85.2} &
  \multicolumn{1}{c|}{\cellcolor[HTML]{C0C0C0}465.0} &
  \multicolumn{1}{c|}{72.7} &
  \multicolumn{1}{c|}{94.8} &
  \multicolumn{1}{c|}{98.4} &
  \multicolumn{1}{c|}{58.8} &
  \multicolumn{1}{c|}{88.4} &
  \multicolumn{1}{c|}{94.8} &
  \multicolumn{1}{c|}{\cellcolor[HTML]{C0C0C0}507.9} &
  \multicolumn{1}{c|}{50.4} &
  \multicolumn{1}{c|}{82.2} &
  \multicolumn{1}{c|}{90.0} &
  \multicolumn{1}{c|}{38.6} &
  \multicolumn{1}{c|}{69.3} &
  \multicolumn{1}{c|}{80.4} &
  \cellcolor[HTML]{C0C0C0}410.9 \\
\multicolumn{1}{l|}{CAMP\_ensemble \cite{wang2019camp} (2019)} &
  \multicolumn{1}{c|}{68.1} &
  \multicolumn{1}{c|}{89.7} &
  \multicolumn{1}{c|}{95.2} &
  \multicolumn{1}{c|}{51.5} &
  \multicolumn{1}{c|}{77.1} &
  \multicolumn{1}{c|}{85.3} &
  \multicolumn{1}{c|}{\cellcolor[HTML]{C0C0C0}466.9} &
  \multicolumn{1}{c|}{72.3} &
  \multicolumn{1}{c|}{94.8} &
  \multicolumn{1}{c|}{98.3} &
  \multicolumn{1}{c|}{58.5} &
  \multicolumn{1}{c|}{87.9} &
  \multicolumn{1}{c|}{95.0} &
  \multicolumn{1}{c|}{\cellcolor[HTML]{C0C0C0}506.8} &
  \multicolumn{1}{c|}{50.1} &
  \multicolumn{1}{c|}{82.1} &
  \multicolumn{1}{c|}{89.7} &
  \multicolumn{1}{c|}{39.0} &
  \multicolumn{1}{c|}{68.9} &
  \multicolumn{1}{c|}{80.2} &
  \cellcolor[HTML]{C0C0C0}410.0 \\
\multicolumn{1}{l|}{CAAN\_ensemble \cite{zhang2020context} (2020)} &
  \multicolumn{1}{c|}{70.1} &
  \multicolumn{1}{c|}{91.6} &
  \multicolumn{1}{c|}{97.2} &
  \multicolumn{1}{c|}{52.8} &
  \multicolumn{1}{c|}{79.0} &
  \multicolumn{1}{c|}{87.9} &
  \multicolumn{1}{c|}{\cellcolor[HTML]{C0C0C0}478.6} &
  \multicolumn{1}{c|}{75.5} &
  \multicolumn{1}{c|}{95.4} &
  \multicolumn{1}{c|}{98.5} &
  \multicolumn{1}{c|}{61.3} &
  \multicolumn{1}{c|}{89.7} &
  \multicolumn{1}{c|}{95.2} &
  \multicolumn{1}{c|}{\cellcolor[HTML]{C0C0C0}515.6} &
  \multicolumn{1}{c|}{52.5} &
  \multicolumn{1}{c|}{83.3} &
  \multicolumn{1}{c|}{90.9} &
  \multicolumn{1}{c|}{41.2} &
  \multicolumn{1}{c|}{70.3} &
  \multicolumn{1}{c|}{82.9} &
  \cellcolor[HTML]{C0C0C0}421.1 \\
\multicolumn{1}{l|}{SGM\_ensemble \cite{wang2020cross} (2020)} &
  \multicolumn{1}{c|}{71.8} &
  \multicolumn{1}{c|}{91.7} &
  \multicolumn{1}{c|}{95.5} &
  \multicolumn{1}{c|}{53.5} &
  \multicolumn{1}{c|}{79.6} &
  \multicolumn{1}{c|}{86.5} &
  \multicolumn{1}{c|}{\cellcolor[HTML]{C0C0C0}478.6} &
  \multicolumn{1}{c|}{73.4} &
  \multicolumn{1}{c|}{93.8} &
  \multicolumn{1}{c|}{97.8} &
  \multicolumn{1}{c|}{57.5} &
  \multicolumn{1}{c|}{87.3} &
  \multicolumn{1}{c|}{94.3} &
  \multicolumn{1}{c|}{\cellcolor[HTML]{C0C0C0}504.1} &
  \multicolumn{1}{c|}{50.0} &
  \multicolumn{1}{c|}{79.3} &
  \multicolumn{1}{c|}{87.9} &
  \multicolumn{1}{c|}{35.3} &
  \multicolumn{1}{c|}{64.9} &
  \multicolumn{1}{c|}{76.5} &
  \cellcolor[HTML]{C0C0C0}393.9 \\
\multicolumn{1}{l|}{PFAN\_ensemble \cite{DBLP:conf/ijcai/WangYQMLLF19} (2019)} &
  \multicolumn{1}{c|}{70.0} &
  \multicolumn{1}{c|}{91.8} &
  \multicolumn{1}{c|}{95.0} &
  \multicolumn{1}{c|}{50.4} &
  \multicolumn{1}{c|}{78.7} &
  \multicolumn{1}{c|}{86.1} &
  \multicolumn{1}{c|}{\cellcolor[HTML]{C0C0C0}472.1} &
  \multicolumn{1}{c|}{76.5} &
  \multicolumn{1}{c|}{{\ul 96.3}} &
  \multicolumn{1}{c|}{{\ul 99.0}} &
  \multicolumn{1}{c|}{61.6} &
  \multicolumn{1}{c|}{89.6} &
  \multicolumn{1}{c|}{95.2} &
  \multicolumn{1}{c|}{\cellcolor[HTML]{C0C0C0}518.2} &
  \multicolumn{1}{c|}{-} &
  \multicolumn{1}{c|}{-} &
  \multicolumn{1}{c|}{-} &
  \multicolumn{1}{c|}{-} &
  \multicolumn{1}{c|}{-} &
  \multicolumn{1}{c|}{-} &
  \cellcolor[HTML]{C0C0C0}- \\
\multicolumn{1}{l|}{IMRAM\_ensemble \cite{chen2020imram} (2020)} &
  \multicolumn{1}{c|}{74.1} &
  \multicolumn{1}{c|}{93.0} &
  \multicolumn{1}{c|}{96.6} &
  \multicolumn{1}{c|}{53.9} &
  \multicolumn{1}{c|}{79.4} &
  \multicolumn{1}{c|}{87.2} &
  \multicolumn{1}{c|}{\cellcolor[HTML]{C0C0C0}484.2} &
  \multicolumn{1}{c|}{76.7} &
  \multicolumn{1}{c|}{95.6} &
  \multicolumn{1}{c|}{98.5} &
  \multicolumn{1}{c|}{61.7} &
  \multicolumn{1}{c|}{89.1} &
  \multicolumn{1}{c|}{95.0} &
  \multicolumn{1}{c|}{\cellcolor[HTML]{C0C0C0}516.6} &
  \multicolumn{1}{c|}{53.7} &
  \multicolumn{1}{c|}{83.2} &
  \multicolumn{1}{c|}{91.0} &
  \multicolumn{1}{c|}{39.6} &
  \multicolumn{1}{c|}{69.1} &
  \multicolumn{1}{c|}{79.8} &
  \cellcolor[HTML]{C0C0C0}416.4 \\
\multicolumn{1}{l|}{ADAPT\_ensemble \cite{wehrmann2020adaptive} (2020)} &
  \multicolumn{1}{c|}{76.6} &
  \multicolumn{1}{c|}{95.4} &
  \multicolumn{1}{c|}{97.6} &
  \multicolumn{1}{c|}{60.7} &
  \multicolumn{1}{c|}{86.6} &
  \multicolumn{1}{c|}{92.0} &
  \multicolumn{1}{c|}{\cellcolor[HTML]{C0C0C0}508.9} &
  \multicolumn{1}{c|}{76.5} &
  \multicolumn{1}{c|}{95.6} &
  \multicolumn{1}{c|}{{98.9}} &
  \multicolumn{1}{c|}{62.2} &
  \multicolumn{1}{c|}{90.5} &
  \multicolumn{1}{c|}{96.0} &
  \multicolumn{1}{c|}{\cellcolor[HTML]{C0C0C0}519.7} &
  \multicolumn{1}{c|}{-} &
  \multicolumn{1}{c|}{-} &
  \multicolumn{1}{c|}{-} &
  \multicolumn{1}{c|}{-} &
  \multicolumn{1}{c|}{-} &
  \multicolumn{1}{c|}{-} &
  \cellcolor[HTML]{C0C0C0}- \\
\multicolumn{1}{l|}{DIME\_ensemble \cite{qu2021dynamic} (2021)} &
  \multicolumn{1}{c|}{{\ul 81.0}} &
  \multicolumn{1}{c|}{{\ul 95.9}} &
  \multicolumn{1}{c|}{{\ul 98.4}} &
  \multicolumn{1}{c|}{{\ul 63.6}} &
  \multicolumn{1}{c|}{{\ul 88.1}} &
  \multicolumn{1}{c|}{{\ul 93.0}} &
  \multicolumn{1}{c|}{\cellcolor[HTML]{C0C0C0}{\ul 520.0}} &
  \multicolumn{1}{c|}{{\ul 78.8}} &
  \multicolumn{1}{c|}{{\ul 96.3}} &
  \multicolumn{1}{c|}{98.7} &
  \multicolumn{1}{c|}{{\ul 64.8}} &
  \multicolumn{1}{c|}{{\ul 91.5}} &
  \multicolumn{1}{c|}{{\ul 96.5}} &
  \multicolumn{1}{c|}{\cellcolor[HTML]{C0C0C0}{\ul 526.6}} &
  \multicolumn{1}{c|}{{\ul 59.3}} &
  \multicolumn{1}{c|}{{\ul 85.4}} &
  \multicolumn{1}{c|}{{\ul 91.9}} &
  \multicolumn{1}{c|}{{\ul 43.1}} &
  \multicolumn{1}{c|}{{\ul 73.0}} &
  \multicolumn{1}{c|}{{\ul 83.1}} &
  \cellcolor[HTML]{C0C0C0}{\ul 435.8} \\ \hline
\multicolumn{22}{c}{\textbf{Ours}} \\ \hline
\multicolumn{1}{l|}{VSRN*} &
  \multicolumn{1}{c|}{69.8} &
  \multicolumn{1}{c|}{88.8} &
  \multicolumn{1}{c|}{93.6} &
  \multicolumn{1}{c|}{52.2} &
  \multicolumn{1}{c|}{78.3} &
  \multicolumn{1}{c|}{86.1} &
  \multicolumn{1}{c|}{\cellcolor[HTML]{C0C0C0}468.8} &
  \multicolumn{1}{c|}{73.8} &
  \multicolumn{1}{c|}{94.1} &
  \multicolumn{1}{c|}{97.9} &
  \multicolumn{1}{c|}{60.1} &
  \multicolumn{1}{c|}{88.1} &
  \multicolumn{1}{c|}{94.0} &
  \multicolumn{1}{c|}{\cellcolor[HTML]{C0C0C0}508.0} &
  \multicolumn{1}{c|}{49.7} &
  \multicolumn{1}{c|}{78.5} &
  \multicolumn{1}{c|}{87.6} &
  \multicolumn{1}{c|}{37.2} &
  \multicolumn{1}{c|}{68.3} &
  \multicolumn{1}{c|}{79.0} &
  \cellcolor[HTML]{C0C0C0}400.3 \\
\multicolumn{1}{l|}{VSRN*+SelfAlign} &
  \multicolumn{1}{c|}{{\color[HTML]{3531FF} \textbf{70.4}}} &
  \multicolumn{1}{c|}{{\color[HTML]{3531FF} \textbf{91.7}}} &
  \multicolumn{1}{c|}{{\color[HTML]{3531FF} \textbf{95.9}}} &
  \multicolumn{1}{c|}{{\color[HTML]{3531FF} \textbf{54.6}}} &
  \multicolumn{1}{c|}{{\color[HTML]{3531FF} \textbf{81.5}}} &
  \multicolumn{1}{c|}{{\color[HTML]{3531FF} \textbf{88.4}}} &
  \multicolumn{1}{c|}{\cellcolor[HTML]{C0C0C0}{\color[HTML]{3531FF} \textbf{482.6}}} &
  \multicolumn{1}{c|}{{\color[HTML]{3531FF} \textbf{74.7}}} &
  \multicolumn{1}{c|}{{\color[HTML]{3531FF} \textbf{95.1}}} &
  \multicolumn{1}{c|}{{\color[HTML]{3531FF} \textbf{98.0}}} &
  \multicolumn{1}{c|}{{\color[HTML]{3531FF} \textbf{62.4}}} &
  \multicolumn{1}{c|}{{\color[HTML]{3531FF} \textbf{89.4}}} &
  \multicolumn{1}{c|}{{\color[HTML]{3531FF} \textbf{95.1}}} &
  \multicolumn{1}{c|}{\cellcolor[HTML]{C0C0C0}{\color[HTML]{3531FF} \textbf{514.7}}} &
  \multicolumn{1}{c|}{{\color[HTML]{3531FF} \textbf{52.4}}} &
  \multicolumn{1}{c|}{{\color[HTML]{3531FF} \textbf{81.3}}} &
  \multicolumn{1}{c|}{{\color[HTML]{3531FF} \textbf{89.3}}} &
  \multicolumn{1}{c|}{{\color[HTML]{3531FF} \textbf{39.6}}} &
  \multicolumn{1}{c|}{{\color[HTML]{3531FF} \textbf{70.0}}} &
  \multicolumn{1}{c|}{{\color[HTML]{3531FF} \textbf{80.7}}} &
  \cellcolor[HTML]{C0C0C0}{\color[HTML]{3531FF} \textbf{413.3}} \\ 
  \multicolumn{1}{l|}{VSRN Improvement} &
  \multicolumn{1}{c|}{{ \textbf{+0.6}}} &
  \multicolumn{1}{c|}{{ \textbf{+2.9}}} &
  \multicolumn{1}{c|}{{ \textbf{+2.3}}} &
  \multicolumn{1}{c|}{{ \textbf{+2.4}}} &
  \multicolumn{1}{c|}{{ \textbf{+3.2}}} &
  \multicolumn{1}{c|}{{ \textbf{+2.3}}} &
  \multicolumn{1}{c|}{\cellcolor[HTML]{C0C0C0}{ \textbf{+13.8}}} &
  \multicolumn{1}{c|}{{ \textbf{+0.9}}} &
  \multicolumn{1}{c|}{{ \textbf{+1.0}}} &
  \multicolumn{1}{c|}{{ \textbf{+0.1}}} &
  \multicolumn{1}{c|}{{ \textbf{+2.3}}} &
  \multicolumn{1}{c|}{{ \textbf{+1.3}}} &
  \multicolumn{1}{c|}{{ \textbf{+1.1}}} &
  \multicolumn{1}{c|}{\cellcolor[HTML]{C0C0C0}{ \textbf{+6.7}}} &
  \multicolumn{1}{c|}{{ \textbf{+2.7}}} &
  \multicolumn{1}{c|}{{ \textbf{+2.8}}} &
  \multicolumn{1}{c|}{{ \textbf{+1.7}}} &
  \multicolumn{1}{c|}{{ \textbf{+2.4}}} &
  \multicolumn{1}{c|}{{ \textbf{+1.7}}} &
  \multicolumn{1}{c|}{{ \textbf{+1.7}}} &
  \cellcolor[HTML]{C0C0C0}{ \textbf{+13.0}} \\ \hline
\multicolumn{1}{l|}{{\color[HTML]{000000} VSRN*\_ensemble}} &
  \multicolumn{1}{c|}{{\color[HTML]{333333} 71.0}} &
  \multicolumn{1}{c|}{{\color[HTML]{333333} 90.6}} &
  \multicolumn{1}{c|}{{\color[HTML]{333333} 94.3}} &
  \multicolumn{1}{c|}{{\color[HTML]{333333} 53.9}} &
  \multicolumn{1}{c|}{{\color[HTML]{333333} 80.3}} &
  \multicolumn{1}{c|}{{\color[HTML]{333333} 87.0}} &
  \multicolumn{1}{c|}{\cellcolor[HTML]{C0C0C0}{\color[HTML]{333333} 477.1}} &
  \multicolumn{1}{c|}{{\color[HTML]{333333} 74.8}} &
  \multicolumn{1}{c|}{{\color[HTML]{333333} 95.1}} &
  \multicolumn{1}{c|}{{\color[HTML]{333333} 98.3}} &
  \multicolumn{1}{c|}{{\color[HTML]{333333} 62.7}} &
  \multicolumn{1}{c|}{{\color[HTML]{333333} 89.8}} &
  \multicolumn{1}{c|}{{\color[HTML]{333333} 95.0}} &
  \multicolumn{1}{c|}{\cellcolor[HTML]{C0C0C0}{\color[HTML]{333333} 515.7}} &
  \multicolumn{1}{c|}{{\color[HTML]{333333} 51.7}} &
  \multicolumn{1}{c|}{{\color[HTML]{333333} 80.8}} &
  \multicolumn{1}{c|}{{\color[HTML]{333333} 88.8}} &
  \multicolumn{1}{c|}{{\color[HTML]{333333} 39.9}} &
  \multicolumn{1}{c|}{{\color[HTML]{333333} 70.4}} &
  \multicolumn{1}{c|}{{\color[HTML]{333333} 81.1}} &
  \cellcolor[HTML]{C0C0C0}{\color[HTML]{333333} 412.7} \\
\multicolumn{1}{l|}{(VSRN*+SelfAlign)\_ensemble} &
  \multicolumn{1}{c|}{{\color[HTML]{3531FF} \textbf{72.2}}} &
  \multicolumn{1}{c|}{{\color[HTML]{3531FF} \textbf{92.8}}} &
  \multicolumn{1}{c|}{{\color[HTML]{3531FF} \textbf{96.6}}} &
  \multicolumn{1}{c|}{{\color[HTML]{3531FF} \textbf{55.8}}} &
  \multicolumn{1}{c|}{{\color[HTML]{3531FF} \textbf{82.7}}} &
  \multicolumn{1}{c|}{{\color[HTML]{3531FF} \textbf{89.0}}} &
  \multicolumn{1}{c|}{\cellcolor[HTML]{C0C0C0}{\color[HTML]{3531FF} \textbf{489.1}}} &
  \multicolumn{1}{c|}{{\color[HTML]{3531FF} \textbf{75.8}}} &
  \multicolumn{1}{c|}{{\color[HTML]{3531FF} \textbf{95.5}}} &
  \multicolumn{1}{c|}{{\color[HTML]{3531FF} \textbf{98.6}}} &
  \multicolumn{1}{c|}{{\color[HTML]{3531FF} \textbf{64.1}}} &
  \multicolumn{1}{c|}{{\color[HTML]{3531FF} \textbf{90.5}}} &
  \multicolumn{1}{c|}{{\color[HTML]{3531FF} \textbf{95.8}}} &
  \multicolumn{1}{c|}{\cellcolor[HTML]{C0C0C0}{\color[HTML]{3531FF} \textbf{520.3}}} &
  \multicolumn{1}{c|}{{\color[HTML]{3531FF} \textbf{54.3}}} &
  \multicolumn{1}{c|}{{\color[HTML]{3531FF} \textbf{82.4}}} &
  \multicolumn{1}{c|}{{\color[HTML]{3531FF} \textbf{90.2}}} &
  \multicolumn{1}{c|}{{\color[HTML]{3531FF} \textbf{41.3}}} &
  \multicolumn{1}{c|}{{\color[HTML]{3531FF} \textbf{71.7}}} &
  \multicolumn{1}{c|}{{\color[HTML]{3531FF} \textbf{82.2}}} &
  \cellcolor[HTML]{C0C0C0}{\color[HTML]{3531FF} \textbf{422.1}} \\ 
  \multicolumn{1}{l|}{ VSRN*\_ensemble Improvement} &
  \multicolumn{1}{c|}{{ \textbf{+1.2}}} &
  \multicolumn{1}{c|}{{ \textbf{+2.2}}} &
  \multicolumn{1}{c|}{{ \textbf{+2.3}}} &
  \multicolumn{1}{c|}{{ \textbf{+1.9}}} &
  \multicolumn{1}{c|}{{ \textbf{+2.4}}} &
  \multicolumn{1}{c|}{{ \textbf{+2.0}}} &
  \multicolumn{1}{c|}{\cellcolor[HTML]{C0C0C0}{ \textbf{+12.0}}} &
  \multicolumn{1}{c|}{{ \textbf{+1.0}}} &
  \multicolumn{1}{c|}{{ \textbf{+0.4}}} &
  \multicolumn{1}{c|}{{ \textbf{+0.3}}} &
  \multicolumn{1}{c|}{{ \textbf{+1.4}}} &
  \multicolumn{1}{c|}{{ \textbf{+0.7}}} &
  \multicolumn{1}{c|}{{ \textbf{+0.8}}} &
  \multicolumn{1}{c|}{\cellcolor[HTML]{C0C0C0}{ \textbf{+4.6}}} &
  \multicolumn{1}{c|}{{ \textbf{+2.6}}} &
  \multicolumn{1}{c|}{{ \textbf{+1.6}}} &
  \multicolumn{1}{c|}{{ \textbf{+1.4}}} &
  \multicolumn{1}{c|}{{ \textbf{+1.4}}} &
  \multicolumn{1}{c|}{{ \textbf{+1.3}}} &
  \multicolumn{1}{c|}{{ \textbf{+1.1}}} &
  \cellcolor[HTML]{C0C0C0}{ \textbf{+9.4}} \\ \hline
\multicolumn{1}{l|}{CAMERA*} &
  \multicolumn{1}{c|}{76.5} &
  \multicolumn{1}{c|}{93.6} &
  \multicolumn{1}{c|}{97.3} &
  \multicolumn{1}{c|}{57.9} &
  \multicolumn{1}{c|}{84.6} &
  \multicolumn{1}{c|}{90.5} &
  \multicolumn{1}{c|}{\cellcolor[HTML]{C0C0C0}500.4} &
  \multicolumn{1}{c|}{74.9} &
  \multicolumn{1}{c|}{95.4} &
  \multicolumn{1}{c|}{98.5} &
  \multicolumn{1}{c|}{62.0} &
  \multicolumn{1}{c|}{89.9} &
  \multicolumn{1}{c|}{95.2} &
  \multicolumn{1}{c|}{\cellcolor[HTML]{C0C0C0}515.9} &
  \multicolumn{1}{c|}{52.4} &
  \multicolumn{1}{c|}{81.7} &
  \multicolumn{1}{c|}{89.9} &
  \multicolumn{1}{c|}{38.8} &
  \multicolumn{1}{c|}{70.1} &
  \multicolumn{1}{c|}{81.4} &
  \cellcolor[HTML]{C0C0C0}414.3 \\
\multicolumn{1}{l|}{CAMERA*+SelfAlign} &
  \multicolumn{1}{c|}{{\color[HTML]{3531FF} \textbf{79.6}}} &
  \multicolumn{1}{c|}{{\color[HTML]{3531FF} \textbf{95.1}}} &
  \multicolumn{1}{c|}{{\color[HTML]{3531FF} {\ul \textbf{97.4}}}} &
  \multicolumn{1}{c|}{{\color[HTML]{3531FF} \textbf{59.7}}} &
  \multicolumn{1}{c|}{{\color[HTML]{3531FF} \textbf{86.2}}} &
  \multicolumn{1}{c|}{{\color[HTML]{3531FF} \textbf{91.5}}} &
  \multicolumn{1}{c|}{\cellcolor[HTML]{C0C0C0}{\color[HTML]{3531FF} \textbf{509.5}}} &
  \multicolumn{1}{c|}{{\color[HTML]{3531FF} \textbf{76.8}}} &
  \multicolumn{1}{c|}{{\color[HTML]{3531FF} \textbf{95.4}}} &
  \multicolumn{1}{c|}{{\color[HTML]{3531FF} \textbf{98.5}}} &
  \multicolumn{1}{c|}{{\color[HTML]{3531FF} \textbf{63.1}}} &
  \multicolumn{1}{c|}{{\color[HTML]{3531FF} \textbf{90.5}}} &
  \multicolumn{1}{c|}{{\color[HTML]{3531FF} \textbf{95.8}}} &
  \multicolumn{1}{c|}{\cellcolor[HTML]{C0C0C0}{\color[HTML]{3531FF} \textbf{520.1}}} &
  \multicolumn{1}{c|}{{\color[HTML]{3531FF} \textbf{54.2}}} &
  \multicolumn{1}{c|}{{\color[HTML]{3531FF} \textbf{82.8}}} &
  \multicolumn{1}{c|}{{\color[HTML]{3531FF} \textbf{90.6}}} &
  \multicolumn{1}{c|}{{\color[HTML]{3531FF} \textbf{40.4}}} &
  \multicolumn{1}{c|}{{\color[HTML]{3531FF} \textbf{71.2}}} &
  \multicolumn{1}{c|}{{\color[HTML]{3531FF} \textbf{81.7}}} &
  \cellcolor[HTML]{C0C0C0}{\color[HTML]{3531FF} \textbf{420.9}} \\ 
  \multicolumn{1}{l|}{ CAMERA* Improvement} &
  \multicolumn{1}{c|}{{ \textbf{+3.1}}} &
  \multicolumn{1}{c|}{{ \textbf{+1.5}}} &
  \multicolumn{1}{c|}{{ {\textbf{+0.1}}}} &
  \multicolumn{1}{c|}{{ \textbf{+1.8}}} &
  \multicolumn{1}{c|}{{ \textbf{+1.6}}} &
  \multicolumn{1}{c|}{{ \textbf{+1.0}}} &
  \multicolumn{1}{c|}{\cellcolor[HTML]{C0C0C0}{ \textbf{+9.1}}} &
  \multicolumn{1}{c|}{{ \textbf{+1.9}}} &
  \multicolumn{1}{c|}{{ \textbf{+0.0}}} &
  \multicolumn{1}{c|}{{ \textbf{+0.0}}} &
  \multicolumn{1}{c|}{{ \textbf{+1.1}}} &
  \multicolumn{1}{c|}{{ \textbf{+0.6}}} &
  \multicolumn{1}{c|}{{ \textbf{+0.6}}} &
  \multicolumn{1}{c|}{\cellcolor[HTML]{C0C0C0}{ \textbf{+4.2}}} &
  \multicolumn{1}{c|}{{ \textbf{+1.8}}} &
  \multicolumn{1}{c|}{{ \textbf{+1.1}}} &
  \multicolumn{1}{c|}{{ \textbf{+0.7}}} &
  \multicolumn{1}{c|}{{ \textbf{+1.6}}} &
  \multicolumn{1}{c|}{{ \textbf{+1.1}}} &
  \multicolumn{1}{c|}{{ \textbf{+0.3}}} &
  \cellcolor[HTML]{C0C0C0}{ \textbf{+6.6}} \\ \hline
\multicolumn{1}{l|}{CAMERA*\_ensemble} &
  \multicolumn{1}{c|}{{\color[HTML]{333333} 78.3}} &
  \multicolumn{1}{c|}{{\color[HTML]{333333} 94.4}} &
  \multicolumn{1}{c|}{{\color[HTML]{333333} {\ul 97.4}}} &
  \multicolumn{1}{c|}{{\color[HTML]{333333} 60.5}} &
  \multicolumn{1}{c|}{{\color[HTML]{333333} 86.1}} &
  \multicolumn{1}{c|}{{\color[HTML]{333333} 91.8}} &
  \multicolumn{1}{c|}{\cellcolor[HTML]{C0C0C0}{\color[HTML]{333333} 508.4}} &
  \multicolumn{1}{c|}{{\color[HTML]{333333} 77.0}} &
  \multicolumn{1}{c|}{{\color[HTML]{333333} {\ul 96.3}}} &
  \multicolumn{1}{c|}{{\color[HTML]{333333} 98.6}} &
  \multicolumn{1}{c|}{{\color[HTML]{333333} 63.6}} &
  \multicolumn{1}{c|}{{\color[HTML]{333333} 90.8}} &
  \multicolumn{1}{c|}{{\color[HTML]{333333} 95.8}} &
  \multicolumn{1}{c|}{\cellcolor[HTML]{C0C0C0}{\color[HTML]{333333} 522.1}} &
  \multicolumn{1}{c|}{{\color[HTML]{333333} 55.2}} &
  \multicolumn{1}{c|}{{\color[HTML]{333333} 83.1}} &
  \multicolumn{1}{c|}{{\color[HTML]{333333} 90.9}} &
  \multicolumn{1}{c|}{{\color[HTML]{333333} 40.4}} &
  \multicolumn{1}{c|}{{\color[HTML]{333333} 71.4}} &
  \multicolumn{1}{c|}{{\color[HTML]{333333} 82.2}} &
  \cellcolor[HTML]{C0C0C0}{\color[HTML]{333333} 423.2} \\
\multicolumn{1}{l|}{(CAMERA*+SelfAlign)\_ensemble} &
  \multicolumn{1}{c|}{{\color[HTML]{3531FF} {\ul \textbf{81.4}}}} &
  \multicolumn{1}{c|}{{\color[HTML]{3531FF} {\ul \textbf{95.6}}}} &
  \multicolumn{1}{c|}{{\color[HTML]{333333} 97.3}} &
  \multicolumn{1}{c|}{{\color[HTML]{3531FF} {\ul \textbf{61.5}}}} &
  \multicolumn{1}{c|}{{\color[HTML]{3531FF} {\ul \textbf{87.1}}}} &
  \multicolumn{1}{c|}{{\color[HTML]{3531FF} {\ul \textbf{92.7}}}} &
  \multicolumn{1}{c|}{\cellcolor[HTML]{C0C0C0}{\color[HTML]{3531FF} {\ul \textbf{515.6}}}} &
  \multicolumn{1}{c|}{{\color[HTML]{3531FF} {\ul \textbf{77.7}}}} &
  \multicolumn{1}{c|}{{\color[HTML]{3531FF} {\ul \textbf{96.3}}}} &
  \multicolumn{1}{c|}{{\color[HTML]{3531FF} {\ul \textbf{98.7}}}} &
  \multicolumn{1}{c|}{{\color[HTML]{3531FF} {\ul \textbf{64.3}}}} &
  \multicolumn{1}{c|}{{\color[HTML]{3531FF} {\ul \textbf{91.3}}}} &
  \multicolumn{1}{c|}{{\color[HTML]{3531FF} {\ul \textbf{96.2}}}} &
  \multicolumn{1}{c|}{\cellcolor[HTML]{C0C0C0}{\color[HTML]{3531FF} {\ul \textbf{524.5}}}} &
  \multicolumn{1}{c|}{{\color[HTML]{3531FF} {\ul \textbf{56.1}}}} &
  \multicolumn{1}{c|}{{\color[HTML]{3531FF} {\ul \textbf{83.6}}}} &
  \multicolumn{1}{c|}{{\color[HTML]{3531FF} {\ul \textbf{91.0}}}} &
  \multicolumn{1}{c|}{{\color[HTML]{3531FF} {\ul \textbf{42.2}}}} &
  \multicolumn{1}{c|}{{\color[HTML]{3531FF} {\ul \textbf{72.5}}}} &
  \multicolumn{1}{c|}{{\color[HTML]{3531FF} {\ul \textbf{82.9}}}} &
  \cellcolor[HTML]{C0C0C0}{\color[HTML]{3531FF} {\ul \textbf{428.3}}} \\
  \multicolumn{1}{l|}{CAMERA*\_ensemble Improvement} &
  \multicolumn{1}{c|}{{ {\textbf{+3.1}}}} &
  \multicolumn{1}{c|}{{ {\textbf{+1.2}}}} &
  \multicolumn{1}{c|}{{\color[HTML]{333333} -0.1}} &
  \multicolumn{1}{c|}{{ {\textbf{+1.0}}}} &
  \multicolumn{1}{c|}{{ {\textbf{+1.0}}}} &
  \multicolumn{1}{c|}{{ {\textbf{+0.9}}}} &
  \multicolumn{1}{c|}{\cellcolor[HTML]{C0C0C0}{ { \textbf{+7.2}}}} &
  \multicolumn{1}{c|}{{ {\textbf{+0.7}}}} &
  \multicolumn{1}{c|}{{ {\textbf{+0.0}}}} &
  \multicolumn{1}{c|}{{ {\textbf{+0.1}}}} &
  \multicolumn{1}{c|}{{ {\textbf{+0.7}}}} &
  \multicolumn{1}{c|}{{ {\textbf{+0.5}}}} &
  \multicolumn{1}{c|}{{ {\textbf{+0.4}}}} &
  \multicolumn{1}{c|}{\cellcolor[HTML]{C0C0C0}{ { \textbf{+2.4}}}} &
  \multicolumn{1}{c|}{{ {\textbf{+0.9}}}} &
  \multicolumn{1}{c|}{{ {\textbf{+0.5}}}} &
  \multicolumn{1}{c|}{{ {\textbf{+0.1}}}} &
  \multicolumn{1}{c|}{{ {\textbf{+1.8}}}} &
  \multicolumn{1}{c|}{{ {\textbf{+1.1}}}} &
  \multicolumn{1}{c|}{{ {\textbf{+0.7}}}} &
  \cellcolor[HTML]{C0C0C0}{ {\textbf{+5.1}}} \\ \hline
\end{tabular}%
 }
\label{sota_cmp}
\end{table*}
To verify the effectiveness of SelfAlign, we compare our results with the state-of-the-art models on Flickr30k and MS-COCO in Table \ref{sota_cmp}.  This table is split into three blocks, from top to bottom, representing independent-embedding models, interactive-embedding models, and our models (\textit{i.e.} VSRN with SelfAlign and CAMERA with SelfAlign), respectively. Notably, for a fair comparison with the interactive-embedding models, following \cite{li2019visual}, we achieve our ensemble models by averaging the predicted similarity scores of the two different models obtained by utilizing different seeds for training.

From the comparison between the first block and last block in Table \ref{sota_cmp}, we conclude that our module SelfAlign can remarkably improve baseline independent-embedding models on all the metrics, which proves the effectiveness of learning fine-grained correspondences of word-object and global-to-local. Specifically, SelfAlign consistently improves the performance of VSRN and CAMERA by 11.2\% and 6.6\% in terms of R@sum on Flickr30k, MS-COCO 1K, and MS-COCO 5K. For the strongest baseline model CAMERA, SelfAlign also achieves 9.1\%, 4.2\%, and 6.6\% boosts in terms of R@sum respectively and achieves a new state-of-the-art performance in independent-embedding models. It is worth noting that SelfAlign helps the strongest baseline model CAMERA achieve 1.1\%$\sim $3.1\% boost in terms of R@1 on both retrieval tasks. We also find the performance improvements of R@1 are almost superior than the improvements of R@5/10, indicating that SelfAlign improves the capability of the baseline model to capture fine-grained discrimination on similar images or texts.

From the comparison between the second block and the last block, we conclude that with the help of SelfAlign rather than cross-attention mechanisms, the performance gap between the interactive-embedding models and the independent-embedding models is reduced by a large margin. Specifically, CAMERA with SelfAlign model outperforms the second-best interactive-embedding model ADAPT \cite{wehrmann2020adaptive} by 6.7\%, 4.8\% in terms of R@sum on Flickr30k, MS-COCO 1K. Compared to the-state-of-art interactive-embedding model DIME \cite{qu2021dynamic}, CAMERA with SelfAlign also achieves comparable performance in terms of R@1 on text retrieval on Flickr30K.

\subsection{Efficiency Comparison}
\label{sec_eff}
\begin{table}[]
\caption{GPU time in early feature encoding stage per query on Flickr30k test set.}
\renewcommand\arraystretch{1.2}
\scalebox{0.73}{%
\begin{tabular}{l|c|cccc}
\hline
\multirow{2}{*}{\textbf{Model}} & \multirow{2}{*}{\textbf{Param.(M)}} & \multicolumn{4}{c}{\textbf{\# of candidates}} \\
                 &       & 1K(ms)            & 10K(ms)           & 100K(ms)          & 1000K(ms)         \\ \hline
PFAN \cite{DBLP:conf/ijcai/WangYQMLLF19}            & 12.8  & 2.5               & 2.5               & 2.5               & 2.5               \\
IMRAM \cite{chen2020imram}          & 21.8  & $4.2\times10^{3}$ & $4.2\times10^{4}$ & $4.2\times10^{5}$ & $4.2\times10^{6}$ \\
DIME \cite{qu2021dynamic}            & \textcolor{black}{116.3}   & $3.0\times10^{3}$ & $3.0\times10^{4}$ & $3.0\times10^{5}$ & $3.0\times10^{6}$ \\ \hline
VSRN \cite{li2019visual}            & 137.7 & 7.3               & 7.3               & 7.3               & 7.3               \\
VSRN+SelfAlign (ours)   & 173.6 & 8.7               & 8.7               & 8.7               & 8.7               \\
CAMERA \cite{qu2020context}          & \textcolor{black}{156.2}  & 25.7              & 25.7              & 25.7     & 25.7              \\
CAMERA+SelfAlign (ours) & 183.5  & 28.2              & 28.2              & 28.2              & 28.2              \\ \hline
\end{tabular}%
}
\label{encoding}
\end{table}

\begin{table}[]
\caption{Time in scoring per query on 100K candidates. Inter. denotes the number of the interactions between words and regions. FLOPs denotes the number of floating-point operations. Dim. denotes the dimension of the similarity calculation vector.}
\renewcommand\arraystretch{1.2}
\scalebox{0.85}{%
\begin{tabular}{l|c|c|c|c}
\hline
\textbf{Model}   & \textbf{Inter.}       & \textbf{FLOPs}        & \textbf{Dim.}         & \textbf{GPU TIME(ms)} \\ \hline
PFAN \cite{DBLP:conf/ijcai/WangYQMLLF19}             & $32\times36$          & $\times1152$          & 1024                  & $3.3\times10^{3}$     \\
IMRAM \cite{chen2020imram}           & \multicolumn{1}{l|}{$32\times36\textcolor{black}{\times3}$} & \multicolumn{1}{l|}{$\textcolor{black}{\times3456}$} & \multicolumn{1}{l|}{1024} & \textcolor{black}{$9.9\times10^{3}$}  \\
DIME \cite{qu2021dynamic}             & $1\times1$            & $\times1$             & 256                   & 1.0                   \\ \hline
VSRN \cite{li2019visual}            & $1\times1$            & $\times1$             & 2048                  & 2.0                   \\
VSRN+SelfAlign (ours)   & $1\times1$            & $\times1$             & 6144                  & 2.3                   \\
CAMERA \cite{qu2020context}          & $1\times1$            & $\times1$             & 2048                  & 2.0                   \\
CAMERA+SelfAlign (ours) & $1\times1$            & $\times1$             & 6144                  & 2.3                   \\ \hline
\end{tabular}%
}
\label{scoring}
\end{table}

To verify the advantage of SelfAlign in keeping the efficiency of independent-embedding models, we construct retrieval latency comparison in the inference phase on baseline models, baseline with SelfAlign models and typical interactive-embedding models. Since the retrieval latency is composed of feature encoding latency and scoring latency, we construct the comparison \textcolor{black}{from} these two aspects as shown in Table \ref{encoding} and Table \ref{scoring} respectively. Since the interactions between the object and word can occur in the feature encoding stage only \cite{wang2019camp, zhang2020context, wehrmann2020adaptive, qu2021dynamic}, the scoring stage only \cite{karpathy2015deep, huang2017instance, lee2018stacked, DBLP:conf/ijcai/WangYQMLLF19} or both stages \cite{chen2020imram, wang2020cross}, we select representative interactive-embedding models from the above three kinds \textcolor{black}{of} methods with comparable accuracy as ours for retrieval latency comparison, e.g., DIME \cite{qu2021dynamic}, PFAN \cite{DBLP:conf/ijcai/WangYQMLLF19}, IMRAM \cite{chen2020imram}.

\textbf{Encoding Latency Comparison.} Table \ref{encoding} shows the online encoding latency given per query. Besides, Table \ref{encoding} also reports the model parameter size which mainly affects the encoding time. Though SelfAlign adds about 36M and 27M parameters and 1$\sim$3ms latency to the baselines, VSRN and CAMERA respectively, the encoding latency of the baseline with SelfAlign models still keeps invariant to the number of candidates, which is much lower than the accuracy comparable interactive-embedding models. This is because SelfAlign preserves the independent feature encoding architecture and does not add any cross-modal interactions in the encoding stage. Therefore, the query embedding can be encoded independently without the interactions with the queried candidates, and the embeddings of the queried candidates can be pre-computed offline without the comsuming of online encoding latency.  In contrast, those models performing cross-modal interactions during encoding stage, like IMRAM and DIME, need to encode each text-image pair, resulting in the encoding latency is linearly related \textcolor{black}{to} the number of candidates.

\textbf{Scoring Latency Comparison.} Table \ref{scoring} shows the scoring latency of per query on 100K candidates. The scoring time is linear with the number of interactions between image embeddings and text embeddings. The results show that SelfAlign increases about 0.3ms latency to baseline models while both baseline models with SelfAlign are still 1000 times faster than the model PFAN and IMRAM, which have cross-modal token-wise interactions during the scoring stage. This is because SelfAlign still keeps the baseline model to perform the computation of similarity scores by using simple similarity calculation operations like dot product.


\subsection{Accuracy and Efficiency Joint Comparison}
\label{sec_joint_cmp}

\begin{figure}  
\centering  
\includegraphics[width=3.5in]{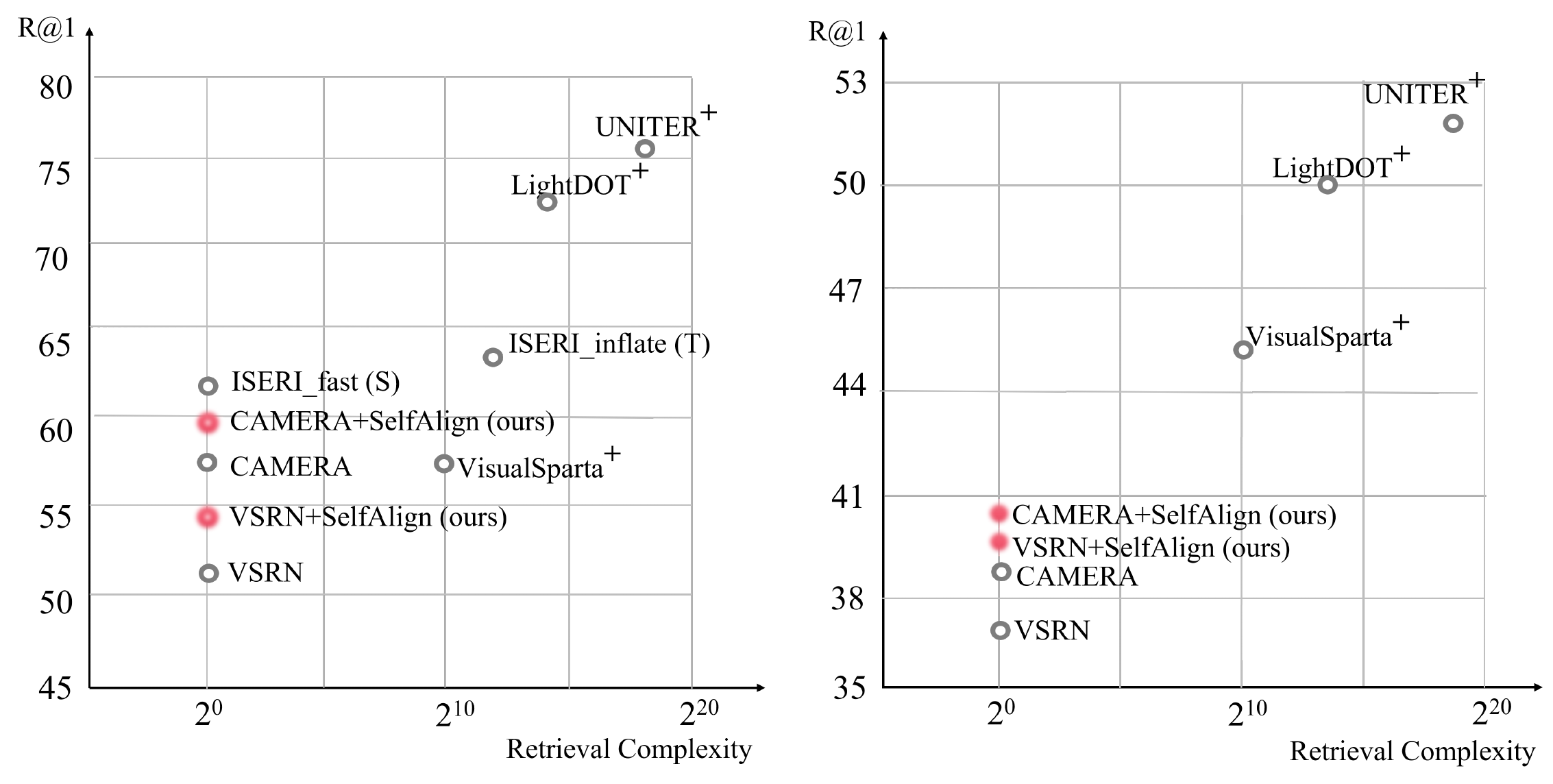}  
\caption{
Image retrieval results of Recall@1 of (a) Flickr30K and (b)  MS-COCO 5K. The retrieval complexity refers to the number of cross-modal interactions included in the early feature encoding stage and late scoring stage. The superscript `+' denotes pre-trained models on huge-scale datasets. \textcolor{black}{`S' and `T' represent the student model and the teacher model respectively.}}  
\label{f30k_joint_cmp}  
\end{figure}

To prove the good balance between accuracy and efficiency for image-text retrieval of our module, we visualize the accuracy and efficiency jointly on Flickr30K and MS-COCO 5K in Figure \ref{f30k_joint_cmp}. Specifically, the trade-off models include VisualSparta \cite{lu2021visualsparta}, ISERI\_inflate \cite{liu2021inflate}, ISERI\_fast \cite{liu2021inflate}, LightDOT \cite{sun2021lightningdot}. VisualSparta \cite{lu2021visualsparta}, ISERI\_inflate \cite{liu2021inflate} are late-interaction trade-off models, where there are no cross-modal token-wise interactions in the feature encoding stage but in the scoring stage. LightDOT \cite{sun2021lightningdot} is a pre-trained \textcolor{black}{re-ranking} based method, which utilizes a pre-trained independent-embedding model to coarsely rank in the first stage and then utilizes another pre-trained interactive-embedding model UNITER \cite{chen2020uniter} to finely rank in the second stage. ISERI\_fast \cite{liu2021inflate} is also an independent-embedding model but is obtained by performing knowledge distillation twice under the extra model supervision of ISERI\_inflate \cite{liu2021inflate}. This enables ISERI\_fast to obtain more accurate fine-grained information under the supervision of the teacher model ISERI\_inflate compared with our models. 

In Figure \ref{f30k_joint_cmp}, the \textit{x}-axis represents the retrieval complexity referring to the average number of cross-modal \textcolor{black}{interactions} per \textcolor{black}{candidate} when given a query to retrieve the ground-truth from 1000 candidates. The number of cross-modal interactions is the sum of the number of interactions in both of the feature encoding stage and the scoring stage. The \textit{y}-axis denotes the results of R@1 in the image retrieval task. The red dots represent our models. As shown in Figure \ref{f30k_joint_cmp}, SelfAlign enables both baseline models improving the accuracy without increasing retrieval complexity while other trade-off models sacrifice retrieval efficiency for cross-modal token-wise interactions. SelfAlign explores the cross-modal semantic correspondences and improves the retrieval efficiency by \textcolor{black}{achieving fine-grained alignment} matching fine-grained correspondences in the early feature encoding stage.

\subsection{Ablation Study}
\label{sec_abl}
\begin{table}[]
\centering
\caption{Ablation study of CAMERA with SelfAlgin on Flick30k.}
\renewcommand\arraystretch{1.1}
\resizebox{0.47\textwidth}{!}{%
\begin{tabular}{ll|c|c|ccc|ccc|
>{\columncolor[HTML]{C0C0C0}}c }
\hline
\multicolumn{2}{l|}{} &
\multicolumn{1}{c|}{\textbf{\textcolor{black}{Param.}}} &
\multicolumn{1}{c|}{\textbf{\textcolor{black}{GPU TIME}}} &
  \multicolumn{3}{c|}{\textbf{Text Retrieval}} &
  \multicolumn{3}{c|}{\textbf{Image Retrieval}} &
  \cellcolor[HTML]{C0C0C0} \\ \cline{5-10}
\multicolumn{2}{l|}{\multirow{-2}{*}{\textbf{Model}}} &
\multicolumn{1}{c|}{\multirow{-1}{*}{\textbf{\textcolor{black}{(M)}}}} &
\multicolumn{1}{c|}{\multirow{-1}{*}{\textbf{\textcolor{black}{(ms)}}}} &
  \textbf{R@1} &
  \textbf{R@5} &
  \textbf{R@10} &
  \textbf{R@1} &
  \textbf{R@5} &
  \textbf{R@10} &
  \multirow{-2}{*}{\cellcolor[HTML]{C0C0C0}\textbf{R@sum}} \\ \hline
\#0       & Full Model        & \textcolor{black}{183.5}  &   \textcolor{black}{2.3}                      & 79.6 & 95.1 & \multicolumn{1}{c|}{97.4} & 59.7 & 86.2 & \multicolumn{1}{c|}{91.5} & 509.5 \\ \hline
\multicolumn{2}{l|}{\textbf{Ablation of Sub-Module}} &
&
&
  \multicolumn{1}{l}{} &
  \multicolumn{1}{l}{} &
  \multicolumn{1}{l|}{} &
  \multicolumn{1}{l}{} &
  \multicolumn{1}{l}{} &
  \multicolumn{1}{l|}{} &
  \multicolumn{1}{l}{\cellcolor[HTML]{C0C0C0}} \\
\#1       & w/o LCA          & \textcolor{black}{183.5}  &  \textcolor{black}{2.3}                       & 76.9 & 94.5 & \multicolumn{1}{c|}{97.5} & 59.6 & 84.8 & \multicolumn{1}{c|}{90.9} & 504.2 \\
\#2       & w/o CRA          & \textcolor{black}{156.2}   & \textcolor{black}{2.0}                       & 77.0 & 95.1 & \multicolumn{1}{c|}{97.2} & 58.7 & 84.6 & \multicolumn{1}{c|}{90.6} & 503.2 \\ \hline
\multicolumn{2}{l|}{\textbf{Ablation of LCA}}  &      &       &      &      & \multicolumn{1}{c|}{}     &      &      & \multicolumn{1}{c|}{}     &       \\
\#3       & w/ Concept O2W   & \textcolor{black}{183.5}     &   \textcolor{black}{2.3}       & 77.7 & 94.3 & \multicolumn{1}{c|}{97.4} & 60.5 & 85.3 & \multicolumn{1}{c|}{91.5} & 506.8 \\
\#4       & w/ Concept Dual  &  \textcolor{black}{183.5}    &   \textcolor{black}{2.3}     & 78.1 & 94.6 & \multicolumn{1}{c|}{97.3} & 60.4 & 85.4 & \multicolumn{1}{c|}{91.3} & 507.1 \\
\#5       & w/ noun+adj+verb   & \textcolor{black}{183.5}    &   \textcolor{black}{2.3}       & 77.2 & 94.2 & \multicolumn{1}{c|}{97.2} & 59.2 & 85.5 & \multicolumn{1}{c|}{91.2} & 504.5 \\
\#6       & w/ noun  &  \textcolor{black}{183.5}    &   \textcolor{black}{2.3}     & 76.4 & 94.8 & \multicolumn{1}{c|}{97.0} & 59.2 & 84.9 & \multicolumn{1}{c|}{91.0} & 503.4 \\ \hline
\multicolumn{2}{l|}{\textbf{Ablation of CRA}}   &      &      &      &      & \multicolumn{1}{c|}{}     &      &      & \multicolumn{1}{c|}{}     &       \\
\#7       &  w/o $\mathcal{L}_{cs}$  & \textcolor{black}{156.2}     &  \textcolor{black}{2.2}      & 76.1 & 94.0 & \multicolumn{1}{c|}{97.9} & 59.4 & 85.3 & \multicolumn{1}{c|}{90.9} & 503.7 \\
\#8       & w/o $\mathcal{L}_{ca}$  &  \textcolor{black}{183.5}    &  \textcolor{black}{2.3}    & 76.1 & 94.6 & \multicolumn{1}{c|}{97.3} & 58.5 & 84.1 & \multicolumn{1}{c|}{89.8} & 500.5 \\
\#9       & w/o $\mathcal{L}_{tg}$ Atd Align  & \textcolor{black}{169.9}    & \textcolor{black}{2.2}     & 77.4 & 95.1 & \multicolumn{1}{c|}{97.4} & 59.4 & 84.7 & \multicolumn{1}{c|}{90.8} & 504.8 \\
\#10       & w/o $\mathcal{L}_{vg}$ Atd Align  &  \textcolor{black}{169.9}    & \textcolor{black}{2.2}     & 78.2 & 94.6 & \multicolumn{1}{c|}{97.4} & 59.3 & 84.8 & \multicolumn{1}{c|}{90.8} & 505.1 \\ \hline
\#11       & CAMERA*                        & \textcolor{black}{156.2}     &  \textcolor{black}{2.0}         & 77.1 & 93.5 & \multicolumn{1}{c|}{96.3} & 58.6 & 84.4 & \multicolumn{1}{c|}{90.6} & 500.5 \\ \hline

\end{tabular}%
\label{ablation_study}
}
\end{table}

We conduct ablation study to evaluate the effectiveness of essential components in SelfAlign. The results on Flickr30k are shown in Table \ref{ablation_study}. We use CAMERA+SelfAlign as the full model for all the following variants: 
\begin{itemize}
\item{\textbf{w/o LCA (model `\#1')}}: this model removes the Local Concept Alignment sub-module in SelfAlign.
\item{\textbf{w/o CRA (model `\#2')}}: this model removes the Contextual Relation Alignment sub-module in SelfAlign.
\item{\textbf{w/ Concept O2W (model `\#3')}}: this model changes the direction of concept alignment learning in LCA sub-module from word-object to object-word. Specifically, we choose the most similar word as the correspondence for each object and perform clustering on word embeddings for fine-grained alignment learning.
\item{\textbf{w/ Concept Dual (model `\#4')}}: this model preserves these two directions of concept alignment learning.
\item{\textbf{w/ noun+adj+verb (model `\#5')}}: \textcolor{black}{this model performs concept alignment learning only with nouns, verbs and adjectives instead of all the words.}
\item{\textbf{w/ noun (model `\#6')}}: \textcolor{black}{this model performs concept alignment learning only with nouns.}
\item{\textbf{w/o $\mathcal{L}_{cs}$ (model `\#7')}}: this model removes the global-to-local contrastive loss $\mathcal{L}_{cs}$ in Equation \ref{loss_cs} for shared context enhancement in CRA sub-module.
\item{\textbf{w/o $\mathcal{L}_{ca}$ (model `\#8')}}: this model removes that the contextual alignment loss $\mathcal{L}_{ca}$ in Equation \ref{loss_ca} for shared context alignment learning in CRA sub-module.
\item{\textbf{w/o $\mathcal{L}_{tg}$ Atd Align (model `\#9')}}: this model removes T-global/V-local contrastive learning attended context alignment. There is no projection for $\bm{t}_{g}^{c}$ in Equation \ref{c_fc},  $\mathcal{L}_{tg}$ in Equation \ref{ctx_tgvl}, text fusion in Equation \ref{fusion} and the similarity computation in the first term of Equation \ref{ctx_sim}.
\item{\textbf{w/o $\mathcal{L}_{vg}$ Atd Align (model `\#10')}}: this model removes V-global/T-local contrastive learning attended context alignment, which is symmetrical with model `\#9'.
\end{itemize}

Models in the first block are designed to evaluate the contribution of each sub-module in SelfAlign. We observe that the R@sum value of models `\#1-\#2' all significantly decrease by 5.3\% and 6.3\%, but they still outperform the baseline model `\#11'. It shows that both alignment sub-modules are effective for baseline to extract different levels of fine-grained correspondences and the combination of them can further improve retrieval accuracy. It proves the effectiveness and \textcolor{black}{complementarity} of concept level and contextual level alignment information learned in LCA and CRA sub-module. 

Models in the second block evaluate the influence of the single direction and \textcolor{black}{the effectiveness of utilizing all words to perform word-object alignment learning} in LCA. The performance of either model `\#3' or model `\#4' decreases slightly by 2.7\% and 2.4\% respectively, which demonstrates word-to-object alignment captures more accurate concept alignment information. 
\textcolor{black}{Besides, The performance of either model `\#5' or model `\#6' decreases by 5.0\% and 6.1\%, which demonstrates the effectiveness of making the most use of text information via utilizing all words to align regions in LCA sub-module.}  

Models in the third block evaluates the influence of the key components in the CRA sub-module. \textcolor{black}{The} performance of model `\#7' and model `\#8' decreases remarkably compared with the full model. The results show that both shared context enhancement by the global-to-local contrastive loss and shared context alignment learning are essential for contextual alignment. The performance of model `\#8' is decreased to the results as the baseline model `\#11' in terms of R@sum, which indicates that CRA sub-module only with global-to-local constrastive learning objective $L_{cs}$ is not only ineffective in context-level alignment but also damaging to the concept-level alignment in the LCA sub-module. The reason is that only with $\mathcal{L}_{cs}$ in the model `\#8' takes the local contextual embeddings from unmatched image-text pairs as the negative samples according to Equation $L_{cs}$ to learn the shared context information in the image-text pair, which ignores that these local contextual embeddings could mainly include the correct concept information in the concept alignment of the LCA sub-module. This conflict makes both the LCA and CRA sub-module ineffective.
Compared to model `\#2', the performance of model `\#9' and model `\#10' improves slightly even when the models only preserve half part of CRA sub-module. These results verify the effectiveness of preserving the joint learning mode of global-to-local contrastive loss. Moreover, compared to the full model, model `\#9' and model `\#10' lead to the performance degradation, which demonstrates that \textcolor{black}{these} two context-level alignment are effective and \textcolor{black}{capture} complementary contextual information from vision global supervision and text global supervision. 

\textcolor{black}{Moreover, we observe that the LCA sub-module does not take extra parameters for the baseline model as it only involves in $\mathcal{L}_{LCA}$ defined in Equation \ref{loc_loss_sum} for local concept alignment learning. The CRA sub-module causes the increments of parameters and retrieval latency for baseline model CAMERA by 27.3M and 0.3ms, respectively, which is due to the feature linear projections defined in Equation \ref{c_fc} and feature fusion based on the gate mechanism defined in Equation \ref{fusion}.}

\subsection{Alignment Quality Analysis}
\begin{figure*} 
\centering 
\includegraphics[width=5.5in]{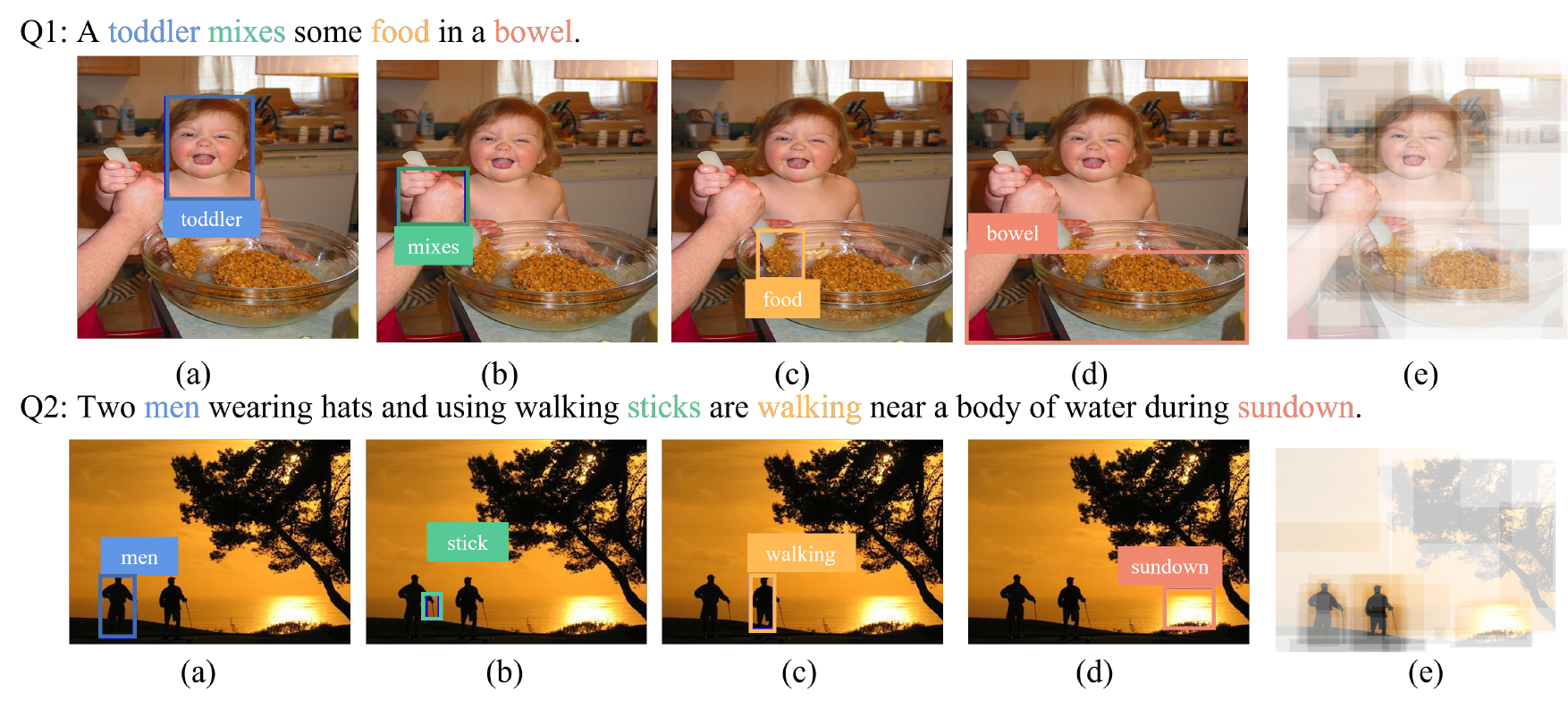}  
\caption{Visualization of similarity in LCA and CRA alignment module. (a) (b) (c) (d) visualize the most matched word-to-region pair results in LCA, where each text concept is used as a query to find the most similar image region and the region outlined in the same color. (e) visualizes the global to local relevance of CRA at the training stage, where the region brightness represents the similarity strength.}
\label{visualization_base_study}
\end{figure*}

\label{sec_vis}
\begin{figure}  
\centering 
\includegraphics[width=3.7in]{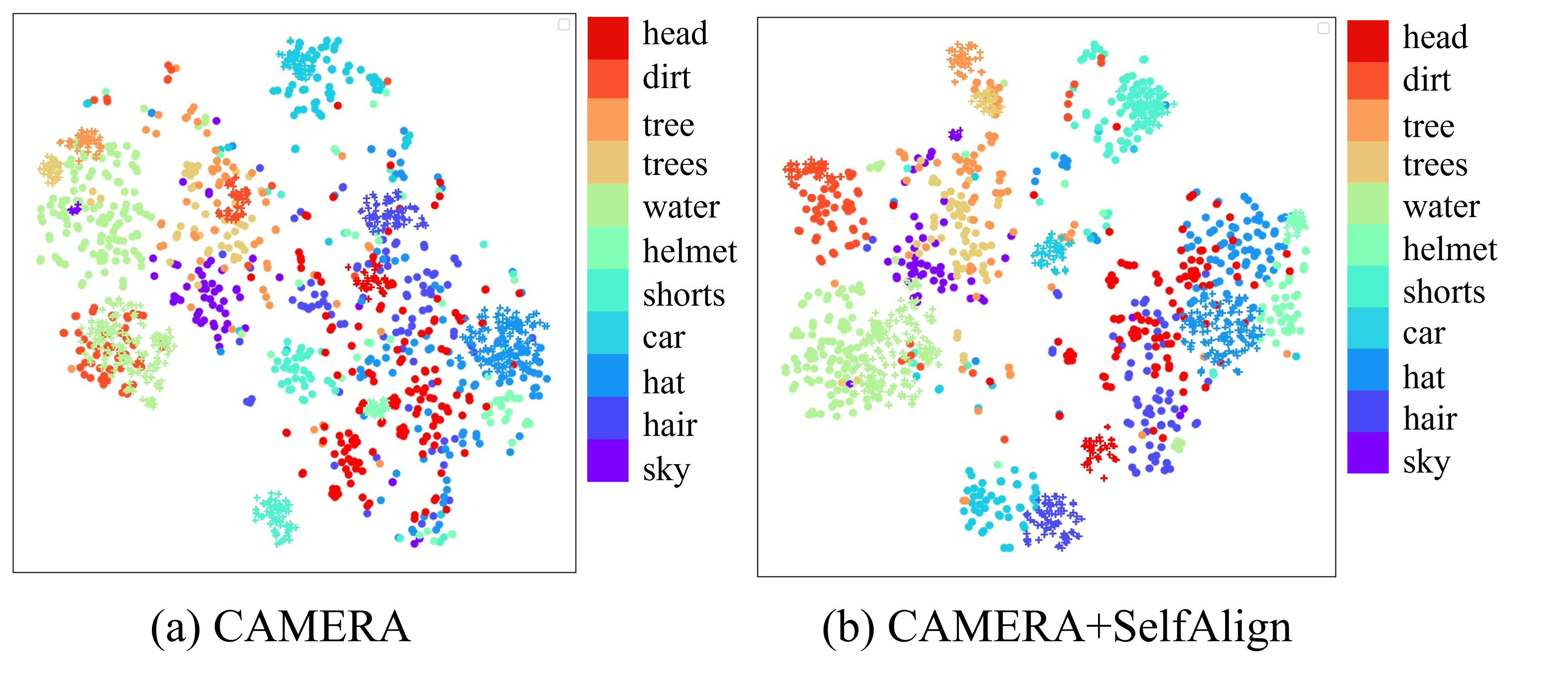}  
\caption{T-SNE visualization of concept representations. `Cross’ means textual concept, and `circle' means visual concept.} 
\label{tnse}
\end{figure}

\textbf{Alignment visualization of the concept representation.} We utilize t-SNE \cite{t-SNE} to visualize region and word embeddings of some frequent concepts for intuitively analyzing the concept-level alignment, as shown in Figure \ref{tnse}. The embeddings are obtained by CAMERA model and CAMERA+SelfAlign model. We conclude that SelfAlign pulls the distance of same semantic concepts of different modalities. For example, in Figure \ref{tnse} (b), the concept `water', `dirt', and `shorts' are densely clustered by CAMERA+SelfAlign while CAMERA can not.   

\textbf{Alignment visualization in LCA and CRA sub-module.} The detailed alignment process is interpretable by visualizing the conceptual and contextual similarity scores at LCA sub-module and CRA sub-module. The results are shown in Figure \ref{visualization_base_study}. Specifically, given an image-text pair in the inference phase, the most similar region for each word is identified in LCA sub-module. The global to local similarity map is computed in CRA sub-module, which supports the explicit visualization and reveals the alignment process. From the first four columns, we observe that the words align to the proper regions with the \textcolor{black}{highest} similarity. For example, in the first example, our model \textcolor{black}{finds} appropriately matched regions on nouns like `toddler', `food', `bowel', as well as the action word like `mixed'. In the CRA sub-module, we observe that semantically matched image regions align well to the text global context embedding, while irrevevant image regions are suppressed. These examples prove that SelfAlign learns the concept-level word-object correspondences in LCA sub-module and composites the essential regions to understand the global context information in CRA.
\begin{figure}  
\centering 
\includegraphics[width=3.5in]{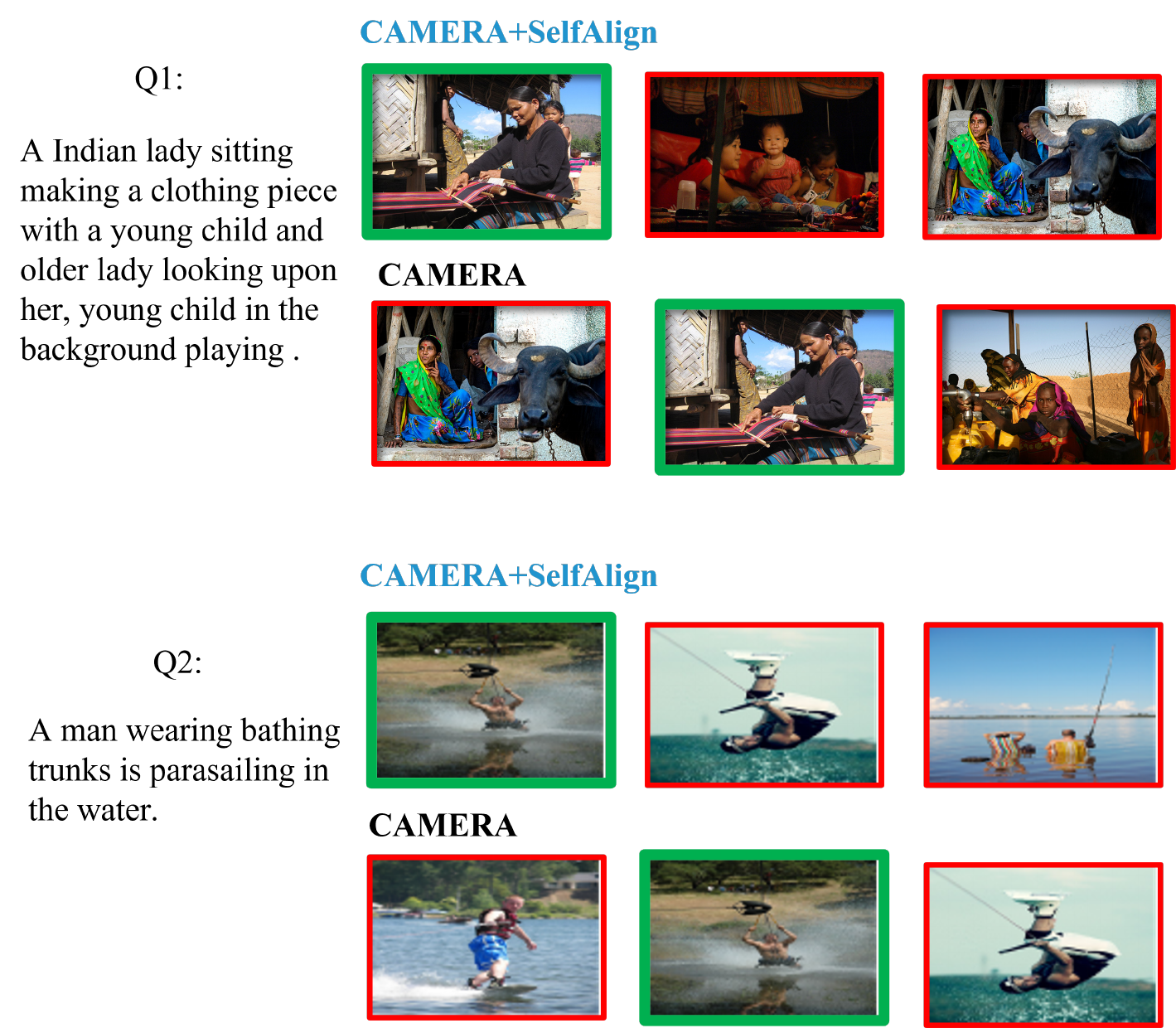}  
\caption{Qualitative comparison of text-to-image retrieval between the baseline model CAMERA and CAMERA+SelfAlign on Flickr30K. We show the top-3 retrieved images for each text query. The truly matched results are marked in green boxes and the falsely matched results are in red boxes.} 
\label{image_retrieval}
\end{figure}
\begin{figure}  
\centering 
\includegraphics[width=3.5in]{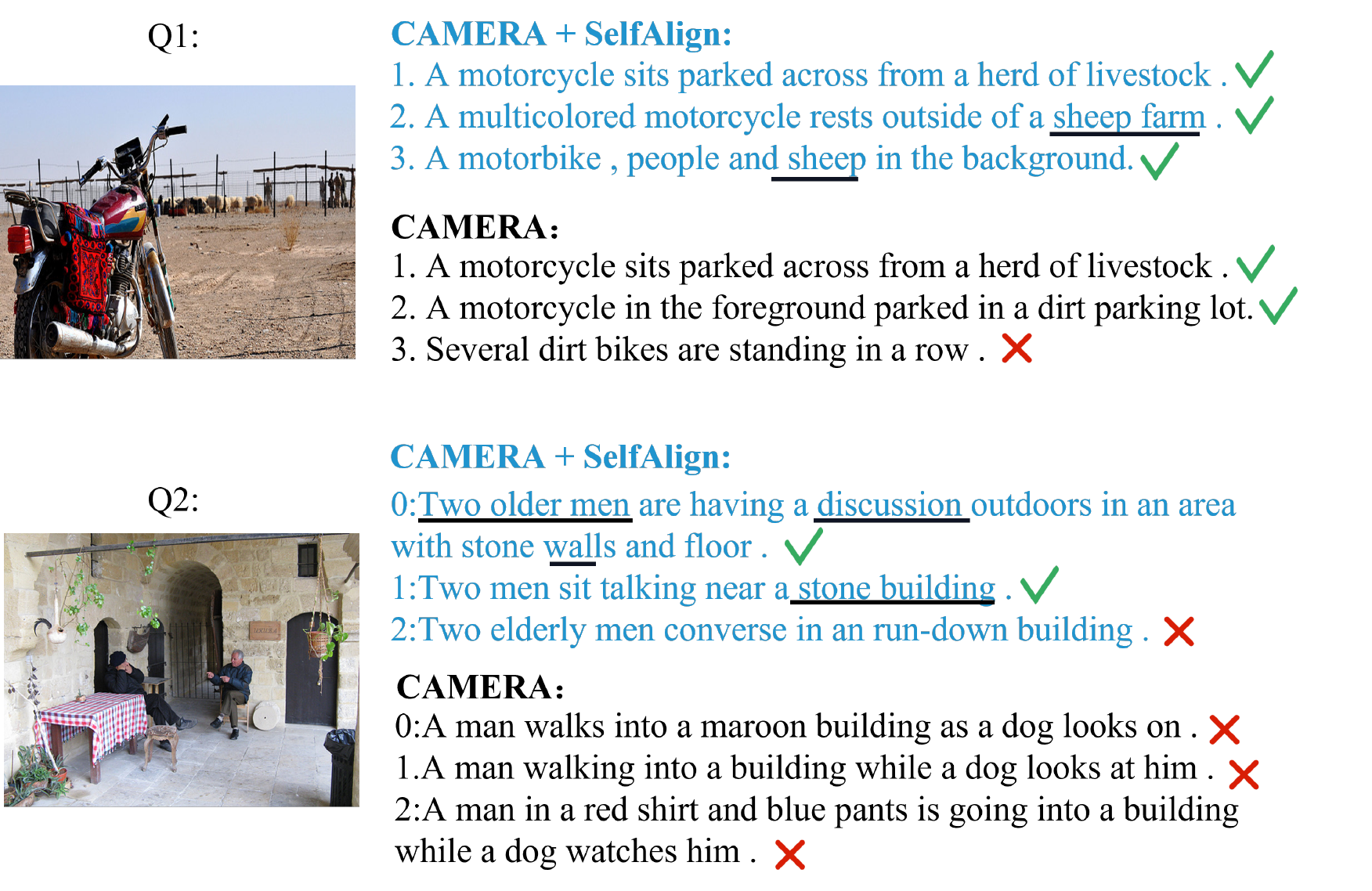}  
\caption{Qualitative comparison of image-to-text retrieval between the baseline model CAMERA and CAMERA+SelfAlign on Flickr30K. We show the top-3 ranked texts for each image query. The truly matched sentences are marked with checks and the falsely matched results are with cross. The concepts in the retrieved text of CAMERA+SelfAlign that differ from the baseline model are marked with underline.} 
\label{text_retrieval}
\end{figure}

\textbf{Qualitative retrieval results analysis.} The qualitative results from text-to-image retrieval and the image-to-text retrieval on Flickr30K are illustrated in Figure \ref{image_retrieval} and Figure \ref{text_retrieval}, respectively. From both retrieved results, it's clear that CAMERA+SelfAlign \textcolor{black}{retrieves} the correct candidates to a more forward position. Moreover, for those candidates with similar scenes, SelfAlign \textcolor{black}{enables} the baseline model to distinguish the fine-grained discrimination among the candidates well, which verifies that SelfAlign is effective in fine-grained cross-modal information retrieval.

\section{Conclusion}
In this paper, we propose a fine-grained image-text alignment module SelfAlign for fast and accurate image-text retrieval. We design two collaborative sub-modules to learn complementary alignment information from both conceptual and contextual level in a self-supervised manner, which improves the retrieval accuracy while keeps the retrieval efficiency. SelfAlign is model-agnostic and generic to \textcolor{black}{incorporate} with various independent-embedding retrieval approaches. Our module consistently boosts the accuracy of the strongest non-pre-training independent-embedding model. How to extend SelfAlign on other cross-modal retrieval \textcolor{black}{tasks} such as video-text retrieval will be our future work.

\bibliographystyle{plainnat}
\footnotesize
\bibliography{ref}

\vspace{-1.2cm}
\begin{IEEEbiography}[{\includegraphics[width=0.8in,height=1in,clip,keepaspectratio]{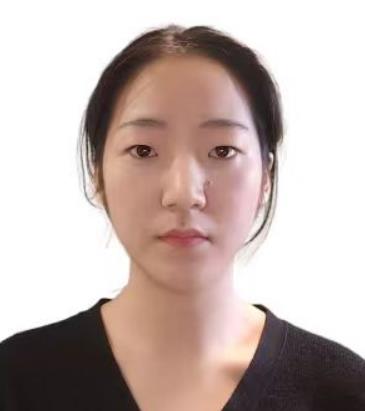}}]{Jiamin Zhuang}
is currently studying for a Ph.D. degree in the Institute of Information Engineering, Chinese Academy of Sciences, Beijing, China. Jiamin Zhuang received her B.S. degree in network engineering from Henan University, China, in 2016. Her research interests mainly focus on cross-modal retrieval.
\end{IEEEbiography}

\vspace{-1.5cm}
\begin{IEEEbiography}[{\includegraphics[width=0.8in,height=1in,clip,keepaspectratio]{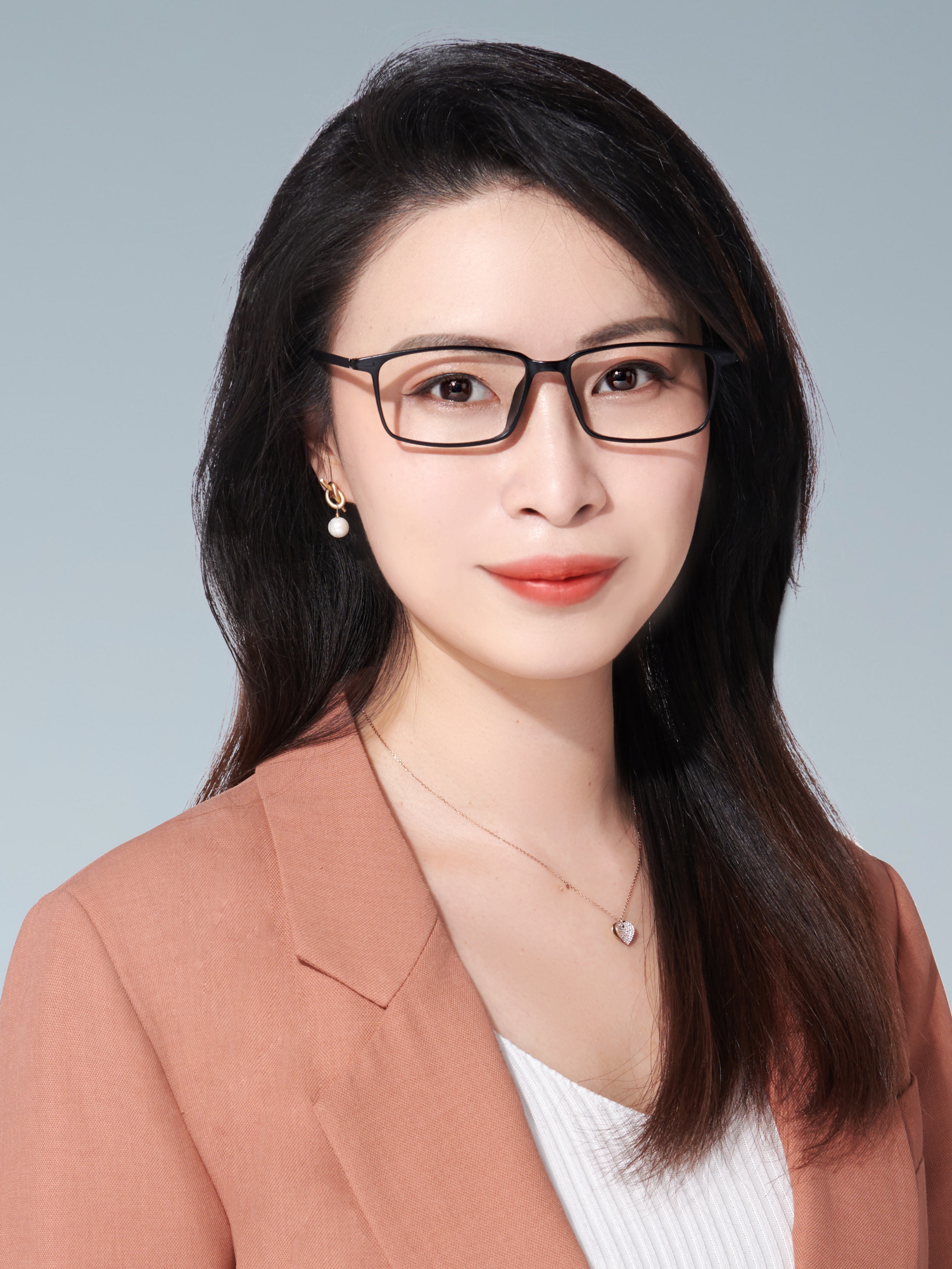}}]{Jing Yu}
is currently an associate professor in the Institute of Information Engineering, Chinese Academy of Sciences, Beijing, China. Jing Yu received her B.S. degree in Automation Science from Minzu University, China, in 2011, and got her M.S. degree in Pattern Recognition from Beihang University, China in 2014. She recieved her Ph.D. degree in the University of Chinese Academy of Sciences, China, in 2019. Her research interests mainly focus on cross-modal understanding, including visual question answering, cross-modal information retrieval, scene graph generation, etc.
\end{IEEEbiography}

\vspace{-1.2cm}
\begin{IEEEbiography}[{\includegraphics[width=0.8in,height=1in,clip,keepaspectratio]{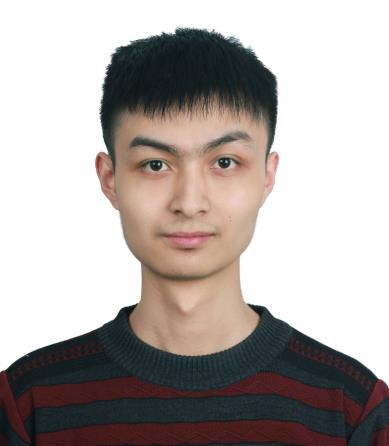}}]{Yang Ding}
is currently studying for a master's degree in the Institute of Information Engineering, Chinese Academy of Sciences, Beijing, China. Yang Ding received his B.S. degree in chemistry from Wuhan University, China, in 2016. His research interests mainly focus on knowledge-based visual question answering.
\end{IEEEbiography}

\vspace{-1.5cm}
\begin{IEEEbiography}[{\includegraphics[width=0.8in,height=1in,clip,keepaspectratio]{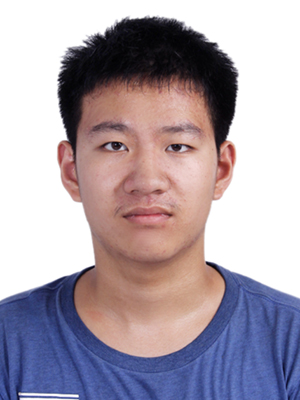}}]{Xiangyan Qu}
is currently studying for a Ph.D. degree in the Institute of Information Engineering, Chinese Academy of Sciences, Beijing, China. Xiangyan Qu received his B.S. degree in Internet of Things Professional from University of Science and Technology Beijing, China, in 2017. His research interests mainly focus on scene recognition.
\end{IEEEbiography}

\vspace{-1.5cm}
\begin{IEEEbiography}[{\includegraphics[width=0.8in,height=1in,clip,keepaspectratio]{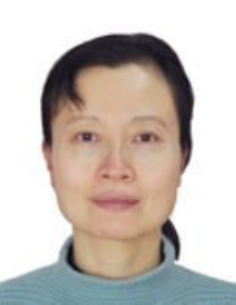}}]{Yue Hu}
is a Professor in the Institute of Information Engineering,Chinese Academy of Sciences, Beijing, China. Her research interests are in the area of natural language processing and social network analysis.
\end{IEEEbiography}

\end{document}